\documentclass{article}

\usepackage[final]{neurips_2024}

\usepackage{cereb}

\usepackage[utf8]{inputenc} %
\usepackage[T1]{fontenc}    %
\PassOptionsToPackage{svgnames,usenames,dvipsnames}{xcolor}
\usepackage{xcolor}
\usepackage[colorlinks=true, citecolor=Navy]{hyperref}
\definecolor{Navy}{RGB}{0,0,128}
\definecolor{ForestGreen}{RGB}{34,139,34}

\usepackage{url}            %
\usepackage{booktabs}       %
\usepackage{amsfonts}       %
\usepackage{nicefrac}       %
\usepackage{microtype}      %

\usepackage{amsmath,amsfonts,bm}

\def\eqref#1{equation~\ref{#1}}

\def\1{\bm{1}}

\DeclareMathAlphabet{\mathsfit}{\encodingdefault}{\sfdefault}{m}{sl}
\SetMathAlphabet{\mathsfit}{bold}{\encodingdefault}{\sfdefault}{bx}{n}

\newcommand{\E}{\mathbb{E}}

\newcommand{\Var}{\mathrm{Var}}

\usepackage{color}
\usepackage{graphicx}
\usepackage{listings}
\usepackage{multirow}
\usepackage{array}
\usepackage{textgreek}
\usepackage{longtable}
\usepackage{caption}
\usepackage{subcaption}
\usepackage{tikz}
\usepackage{wrapfig}
\usepackage{tcolorbox}
\usepackage{enumitem}
\usepackage{rotating}

\setcitestyle{numbers,square,comma}

\usepackage{shortbold}
\usepackage{shortcal}

\usepackage{acronym}
\acrodef{mup}[\textmu P]{maximal update parameterization}
\acrodef{smup}[S\textmu Par]{sparse maximal update parameterization}

\usetikzlibrary{shapes,arrows.meta,positioning,fit,backgrounds,calc}

\newcommand{\muTransfer}{\textmu Transfer}

\newcolumntype{C}[1]{>{\centering\let\newline\\\arraybackslash\hspace{0pt}}m{#1}}
\newcolumntype{L}[1]{>{\raggedright\let\newline\\\arraybackslash\hspace{0pt}}m{#1}}
\newcolumntype{R}[1]{>{\raggedleft\let\newline\\\arraybackslash\hspace{0pt}}m{#1}}

\definecolor{deepblue}{rgb}{0,0,0.7}
\definecolor{deepred}{rgb}{0.7,0,0}
\definecolor{deepgreen}{rgb}{0,0.5,0}

\newcommand\pythonstyle{\lstset{
    language=Python,
    basicstyle=\ttfamily,
    morekeywords={},              %
    keywordstyle=\ttfamily\color{deepblue},
    emph={get_n_params,get_flops_per_seq}, %
    emphstyle=\ttfamily\color{deepred},    %
    commentstyle=\color{deepgreen},
    stringstyle=\color{deepgreen},
    frame=tb,                         %
    showstringspaces=false
}}

\lstnewenvironment{python}[1][] {
    \pythonstyle
    \lstset{#1}
} {}

\makeatletter
\newcommand{\thickhline}{%
    \noalign {\ifnum 0=`}\fi \hrule height 1pt
    \futurelet \reserved@a \@xhline
}
\makeatother

\newcounter{fcounter}
\newcommand\result[1]{
        \refstepcounter{fcounter}\vspace{2pt}
        \begin{tcolorbox}[colback=orange!20!white,colframe=orange!80!black,boxsep=1pt,left=2pt,right=2pt,top=1pt,bottom=1pt]\noindent\emph{\textbf{Finding \arabic{fcounter}}: #1}
        \end{tcolorbox}\vspace{0pt}}

\newcommand\desiderata[1]{
        \begin{tcolorbox}[colback=blue!10!white,colframe=blue!50!black, sharp corners,boxsep=1pt,left=2pt,right=2pt,top=1pt,bottom=1pt]\noindent{#1}
        \end{tcolorbox}\vspace{0pt}}

\title{Sparse maximal update parameterization: A holistic approach to sparse training dynamics}

\author{%
    Nolan Dey \quad
    Shane Bergsma \quad
    Joel Hestness \\
    Cerebras Systems \\
    \texttt{\{nolan,joel\}@cerebras.net}
}

\def\paperyear{2024}

\begin{document}

\maketitle

\begin{abstract}
Several challenges make it difficult for sparse neural networks to compete with dense models.  First, setting a large fraction of weights to zero impairs forward and gradient signal propagation. Second, sparse studies often need to test multiple sparsity levels, while also introducing new hyperparameters (HPs), leading to prohibitive tuning costs. Indeed, the standard practice is to re-use the learning HPs originally crafted for dense models.  Unfortunately, we show sparse and
dense networks do not share the same optimal HPs. Without stable dynamics and effective training recipes, it is costly to test sparsity at scale, which is key to surpassing dense networks and making the business case for sparsity acceleration in hardware.
A holistic approach is needed to tackle these challenges and we propose \ac{smup} as one such approach. For random unstructured static sparsity, \ac{smup} ensures activations, gradients, and weight updates all scale independently of sparsity level. Further, by reparameterizing the HPs, \ac{smup} enables the same HP values to be optimal as we vary both sparsity level and model width. HPs can be tuned on small dense networks and transferred to large sparse models, greatly reducing tuning costs. On large-scale language modeling, \ac{smup} shows increasing improvements over standard parameterization as sparsity increases, leading up to 11.9\% relative loss improvement at 99.2\% sparsity. A minimal implementation of \ac{smup} is available at \url{https://github.com/EleutherAI/nanoGPT-mup/tree/supar}.

\end{abstract}

\section{Intro}\label{sec:intro}

\emph{Sparsity} has emerged as a key technique to mitigate the
increasing computational costs of training and inference in deep
neural networks.
This work focuses on \emph{weight sparsity}, whereby a significant fraction of model weights are kept at zero.
It has long been known that dense neural networks can be heavily pruned 
\emph{after} training~\cite{lecun1989optimal}.
With the goal of reducing costs \emph{during} training,
recent work has explored static weight sparsity from initialization. In this work we focus on random unstructured static sparsity, which has re-emerged as a surprisingly effective strategy~\cite{liu2022unreasonable,thangarasa2023spdf}. This type of sparsity can be accelerated by CPUs, Cerebras, Graphcore, and SambaNova. Furthermore, GPUs and TPUs support 2:4 block structured sparsity which is quite similar to 50\% unstructured sparsity.

Unfortunately, several challenges have hindered progress in weight-sparse neural
networks.  First, sparsity impairs signal propagation during
training~\cite{lee2019signal,evci2022gradient,bambhaniya2024progressive}.
Second, with today's techniques, sparse training is costly.
Sparse techniques typically introduce extra hyperparameters (HPs), e.g., number of pruning iterations at
initialization~\cite{verdenius2020pruning,de2020progressive,tanaka2020pruning},
and it is common to train models across different sparsity levels.
Since tuning should be performed at each level and the search space grows exponentially with the number of HPs,
the tuning costs essentially ``defeat the purpose'' of
sparsity, i.e., to \emph{reduce} computation~\cite{verdenius2020pruning}.
Finally, today there is only a nascent ecosystem of hardware acceleration for unstructured sparsity, so most researchers get little sparsity benefit when tuning.

\begin{figure}[t]
    \centering
    \begin{minipage}{0.57\textwidth}
        \centering
        \begin{tikzpicture}[
            title/.style={font=\sffamily\bfseries}
            ]
            \node[anchor=south west,inner sep=0] (image) at (0,0) {
                \includegraphics[width=\linewidth]{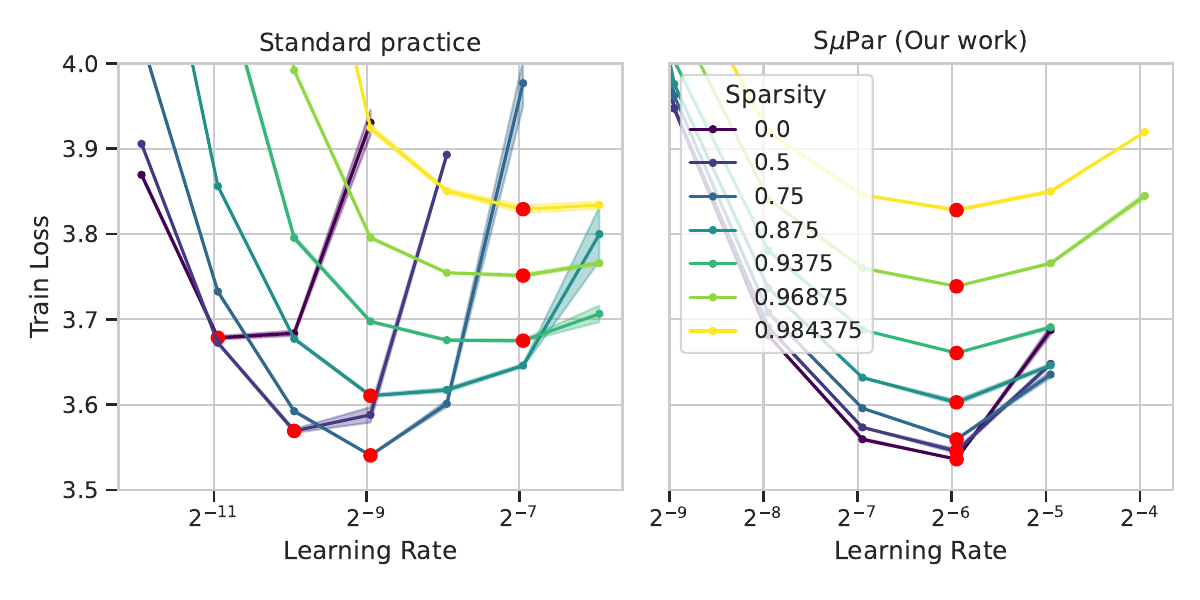}
            };
            \begin{scope}[x={(image.south east)},y={(image.north west)}]
                \draw[-latex, thick, black] (0.45,0.26) -- (0.41,0.32) node[midway, above] (arr1) {};
                \node[title, below=0.05cm of arr1] {\tiny optimum shifts};
                \draw[-latex, thick, black] (0.63,0.26) -- (0.67,0.32) node[midway, above] (arr1) {};
                \node[title, below=0.05cm of arr1] {\tiny optimum stable};
            \end{scope}
        \end{tikzpicture}
        \caption{\ac{smup} (Our work) allows stable optimum HPs for any sparsity level, unlike standard practice.}
        \label{fig:stable_hps}
    \end{minipage}\hfill
    \begin{minipage}{0.4\textwidth}
      \centering
      \scalebox{0.64}{
        {\begin{tikzpicture}[
    node distance=0.2cm and 0.2cm,
    auto,
    block/.style={
        rectangle, draw,
        text width=3.9cm, align=center, minimum height=1cm,
        font=\sffamily\small
    },
    title/.style={
        font=\sffamily\bfseries
    },
    dottedarr/.style={-latex, dotted} %
]

\node[block] (A) {Prohibitive tuning};
\node[block, below=of A] (B) {Inconclusive results};
\node[block, below=of B] (C) {Unclear path to scale};
\node[block, below=0.4cm of C] (D) {Dense still dominant};

\node[title, above=0.15cm of A] () {Standard practice};

\node[block, right=0.5cm of A] (X) {Cheap tuning};
\node[block, below=of X] (Y) {Robust to width/density};
\node[block, below=of Y] (Z) {Ready for next-level scale};
\node[block, below=0.4cm of Z] (W) {Ready to surpass dense};

\node[title, above=0.15cm of X] () {\ac{smup} approach};

\draw[-latex] (A) -- (B);
\draw[-latex] (B) -- (C);
\draw[dottedarr] (C) -- (D);

\draw[-latex] (X) -- (Y);
\draw[-latex] (Y) -- (Z);
\draw[dottedarr] (Z) -- (W);

\end{tikzpicture}}
      }
      \caption{\ac{smup} enables sparse training at scale, helping to surpass dense and motivate sparsity in hardware.}
      \label{fig:flowchart1}
    \end{minipage}
\end{figure}

These costs have led to the standard practice of \emph{simply re-using HPs that were previously optimized for the baseline dense models} (Section~\ref{sec:relwork}).
One might hope that sparse models thrive with the same learning rates
and other HPs as their dense counterparts. Unfortunately, they do not: optimal HPs \emph{systematically} vary with sparsity level (Figure~\ref{fig:stable_hps}, left).
With impaired training dynamics, prohibitive tuning cost, and
lacking the established training recipes enjoyed by dense models,
it is often inefficient to train sparse networks at scale (Figure~\ref{fig:flowchart1}).

To remedy this situation, we propose sparse maximal update parameterization (\ac{smup}, pronounced ``soo-pahr''), a novel, holistic approach to stabilize sparse training dynamics.
\ac{smup} fulfills the Feature Learning Desiderata (Section~\ref{sec:smup}) by parameterizing weight initialization and learning rates with respect to change in width \emph{and} sparsity level. As a generalization of \ac{mup}~\cite{yang2020feature,yang2022mup}, \ac{smup} enjoys well-controlled activation, gradient, and weight update scales in expectation, avoiding exploding or vanishing signal when changing both sparsity and model width.

\begin{wrapfigure}{r}{0.4\textwidth}
    \vspace{-14pt}
    \centering
    \begin{tikzpicture}[
        title/.style={dashed, dash pattern=on 1pt off 1pt, draw, fill=green!10,font=\sffamily\bfseries, align=center, inner xsep=1.5pt, inner ysep=1.5pt},
        myarr/.style={dashed, dash pattern=on 1pt off 1pt, -latex, black}
        ]
        \node[anchor=south west,inner sep=0] (image) at (0,0) {
            \includegraphics[width=0.95\linewidth]{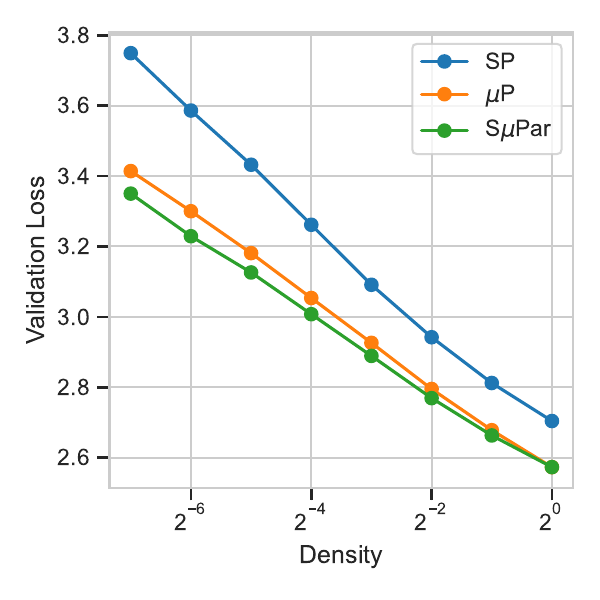}
        };
    \end{tikzpicture}
    \caption{For LLMs, \ac{smup} forms the Pareto frontier loss across sparsity levels, with no HP tuning required.}
    \label{fig:demo-reduced-dense-20tpp}
    \vspace{-6pt}
\end{wrapfigure}

By reparameterizing HPs in this way, \ac{smup} enables the
same HP values to be optimal as sparsity varies 
(Figure~\ref{fig:stable_hps}, right).  We therefore enjoy \muTransfer: we can tune small proxy models and transfer optimal HPs directly to models at scale.
In fact, we discovered our \ac{mup} HPs, tuned for dense models in prior
work (and equivalent to \ac{smup} with sparsity=0\%), correspond to the optimal
learning rate and initial weight variance for \emph{all} sparse models tuned in this paper!
As sparsity increases, our formulation shows the standard parameterization (SP) and \ac{mup} suffer from vanishing signal,
further clarifying prior observations of gradient flow issues in sparse networks.
The improvements enabled by \ac{smup} set the Pareto-frontier best loss across sparsity levels. Figure~\ref{fig:demo-reduced-dense-20tpp} previews this improvement for large language models trained from compute-optimal configurations~\cite{hoffmann2022chinchilla}. Here, \ac{smup} benefits grow with increasing sparsity, to 11.9\% better loss than SP and 1.9\% better loss than \ac{mup} at 99.2\% random unstructured sparsity. See Section \ref{sec:demo} for details on this experiment.

\section{Related work}\label{sec:relwork}

\paragraph{Sparse training landscape}

Sparse training can be divided into static sparsity, where the
connectivity is fixed (our focus) and dynamic sparsity, where the
sparsity mask can evolve~\cite{hoefler2021sparsity}.
We use \emph{unstructured} sparsity, though our approach
generalizes to structured approaches where a particular sparsity
pattern increases efficiency on specific
hardware~\cite{yao2019balanced,kang2019accelerator,mishra2021accelerating,frantar2023sparsegpt,lasby2023dynamic,bambhaniya2024progressive}.
Unstructured connectivity may be based on both
random
pruning~\cite{mocanu2016topological,golubeva2020wider,tessera2021keep,liu2022unreasonable,thangarasa2023spdf}
and various pruning-at-initialization
criteria~\cite{lee2018snip,verdenius2020pruning,wang2020picking,tanaka2020pruning,de2020progressive}.
\citet{liu2022unreasonable} found that as models scale, the relative
performance of randomly pruned networks grow.  Furthermore,
\citet{frantar2023scaling} found the optimal level of sparsity
increases with the amount of training data.
Together, these findings suggest that as neural networks continue to
get wider and deeper, and trained on more and more data, very sparse
randomly-pruned networks may emerge as an attractive option.

\paragraph{Improving sparse training dynamics}

Many prior works identify various sparse training dynamics issues.
In particular, prior works note sparsity impacts weight initialization~\cite{liu2018rethinking,lee2019signal,ramanujan2020what,evci2022gradient},
activation variance~\cite{lasby2023dynamic},
gradient flow~\cite{wang2020picking,lubana2020gradient,tessera2021keep,evci2022gradient,bambhaniya2024progressive},
and step sizes during weight updates~\cite{frantar2023scaling}.
These prior works each only address a subset of these issues in targeted ways,
often showing benefits to sparse model training loss.
We advocate for a holistic approach,
and discuss the relationship between these prior works and our approach in Section~\ref{sec:discussion} after describing and evaluating \ac{smup}.

\paragraph{Sparse sensitivity to HPs}

Due to the costs of training with fixed weight sparsity, re-using
dense HPs is standard practice. Such re-use is typically
indicated in appendices or supplemental materials,
e.g.,~\cite{mocanu2016topological,lee2018snip,liu2018rethinking,lee2019signal,gale2019state,verdenius2020pruning,wang2020picking,tanaka2020pruning,frankle2020pruning,de2020progressive,golubeva2020wider,tessera2021keep,liu2022unreasonable,thangarasa2023spdf}.
Also, dynamic sparsity approaches often compare to
fixed sparsity; these baselines are likewise reported to
re-use the dense
HPs~\cite{bellec2017deep,mocanu2018scalable,evci2020rigging,liu2021we,evci2022gradient,thangarasa2023sparse}.
However, some prior work has suggested such training is sensitive to
HPs, e.g., learning rates~\cite{liu2018rethinking,tessera2021keep},
learning rate schedules~\cite{gale2019state}, or training length \cite{kuznedelev2023accurate}, although systematic
tuning was not performed.
For dynamic sparse training (DST), it is also conventional to re-use
dense HPs, whether in
dense-to-sparse~\cite{lubana2020gradient,frantar2023scaling} or
sparse-to-sparse (evolving mask)
training~\cite{bellec2017deep,dettmers2019sparse,liu2021we,evci2022gradient,thangarasa2023sparse}.
As with fixed sparsity, work here has also suggested sensitivity to
HPs, e.g., to dropout and label smoothing~\cite{gale2019state}.  DST
may also benefit from extra training steps~\cite{evci2020rigging} or
smaller batch sizes~\cite{liu2021we}, although in DST this may mainly be due
to a greater number of opportunities for connectivity
exploration~\cite{liu2021we}.

\section{Sparse maximal update parameterization (S\textmu Par)}\label{sec:smup}
We now provide background, motivation, and derivation for \ac{smup}, first introducing notation (Section \ref{sec:notation}) and then defining Feature Learning Desiderata (Section \ref{sec:fld}) with a brief overview of \ac{mup} (Section \ref{sec:mup}). Finally we motivate \ac{smup} and provide an overview of the parameterization (Section \ref{sec:smup-overview}).

\subsection{Notation} \label{sec:notation}

The operations for a single sparse training step are illustrated in Figure \ref{fig:training-step}. 
The definition and dimensions are:
batch size $B$,
learning rate $\eta$,
loss function $\mathcal{L}$,
forward pass function $\mathcal{F}$,
input dimension $d_\text{in}$,
input activations $\XB \in \mathbb{R}^{B \times d_\text{in}}$,
input activation gradient $\frac{\partial \mathcal{L}}{\partial \XB} = \nabla_{\XB}\mathcal{L} \in \mathbb{R}^{B \times d_\text{in}}$,
output dimension $d_\text{out}$,
output activations $\YB \in \mathbb{R}^{B \times d_\text{out}}$,
output activation gradient $\frac{\partial \mathcal{L}}{\partial \YB} = \nabla_{\YB}\mathcal{L} \in \mathbb{R}^{B \times d_\text{out}}$,
weights $\WB \in \mathbb{R}^{d_\text{in} \times d_\text{out}}$,
initialization variance $\sigma_{W}$ for weights $\WB$,
weight update $\Delta \WB \in \mathbb{R}^{d_\text{in} \times d_\text{out}}$,
and $\Delta \YB \in \mathbb{R}^{B \times d_\text{out}}$ is the effect of the weight update on output activations: $\Delta \YB = \XB (\Delta \WB \odot \MB)$. Unless otherwise specified, $\MB \in \{0,1 \}^{d_\text{in} \times d_\text{out}}$ is an unstructured random static mask with sparsity $s$ and density $\rho = 1-s$. When changing model scale or sparsity, we refer to a width multiplier $m_{d} = \frac{d_\text{in}}{d_\text{in, base}} = \frac{d_\text{out}}{d_\text{out, base}}$ and density multiplier $m_{\rho} = \frac{\rho}{\rho_{\text{base}}}$.

\begin{figure}[h]
    \centering
    \scalebox{0.75}{
      {\begin{tikzpicture}[
    node distance=0.6cm and 0.8cm,
    auto,
    block/.style={rectangle, draw, fill=pink!100, text width=5em, align=center, minimum height=2em},
    box2/.style={rectangle, draw, fill=pink!50, text width=5em, align=center, minimum height=2em},
    circ/.style={circle, draw, fill=yellow!30, minimum size=2em, font=\sffamily\bfseries},
    plus/.style={circle, draw, fill=white!20, minimum size=2em, font=\sffamily\bfseries},
    opt_ellipse/.style={ellipse, draw, fill=yellow!30, minimum width=5em, minimum height=3em, font=\sffamily\bfseries},
    input/.style={rectangle, draw, fill=gray!20, text width=3.7em, align=center, minimum height=2em, font=\sffamily\bfseries},
    grad/.style={rectangle, draw, fill=blue!20, text width=3.7em, align=center, minimum height=2em},
    line/.style={draw, -Latex},
    mydash/.style={dashed, draw, inner xsep=8pt, inner ysep=7pt, dash pattern=on 3pt off 1pt, rounded corners=5pt},
    mydashb/.style={dashed, draw, inner xsep=8pt, inner ysep=6pt, dash pattern=on 3pt off 1pt, rounded corners=5pt},
    dashedline/.style={dashed, line, dash pattern=on 3pt off 1pt, line width=0.15mm},
    heavyline/.style={line,line width=0.2mm}
]

\node[input] (xl) {$\XB$};
\node[circ, right=of xl] (F) {$\mathcal{F}$};
\node[block, below=of F] (wml) {$\WB \odot \MB$};
\node[circ, below=of wml] (back) {$\partial \mathcal{F}$};
\node[grad, left=of back] (vxl) {$\nabla_{\XB} \mathcal{L}$};
\node[plus, right=of wml] (plus) {+};
\node[box2, right=of plus] (deltaW) {$\Delta \WB \odot \MB$};
\node[opt_ellipse, right=of deltaW] (opt) {Optimizer};
\node[grad, right=of opt] (vxl1) {$\nabla_{\YB} \mathcal{L}$};
\node[input] at ($(xl)!(vxl1)!(F)$) (xl1) {$\YB$};

\draw[heavyline] (xl) -- (F);
\draw[heavyline] (F) -- (xl1);
\draw[heavyline] (plus) -- (wml);
\draw[heavyline] (deltaW) -- (plus);
\draw[heavyline] (opt) -- (deltaW);
\draw[heavyline] (vxl1) -- (opt);
\draw[heavyline] (vxl1) |- (back);
\draw[heavyline] (back) -- (vxl);

\draw[heavyline] (wml) -- (F);
\draw[heavyline] (wml) -- (back);

\draw[heavyline]
  (xl)
  -- ++(0,1.2) %
  -| (opt.north); %
\draw[heavyline] (wml) -- ++(0,0.6) -| (plus);

\draw[dashedline] ([xshift=-0.65cm]xl.west) -- (xl.west);
\draw[dashedline] (xl1.east) -- ([xshift=0.65cm]xl1.east);
\draw[dashedline] (vxl.west) -- ([xshift=-0.65cm]vxl.west);
\draw[dashedline] ([xshift=0.65cm]vxl1.east) -- (vxl1.east);

\node[mydash, fit=(F), label=above:Forward] {};
\node[mydash, fit=(back), label=below:Backward] (backwardlabel) {};
\node[mydash, fit=(opt) (plus), label=below:Weight Update] {};
\coordinate (Top) at ([yshift=0.8cm]xl1.north); %
\coordinate (Bottom) at ([yshift=-0.7cm]vxl.south); %
\node[mydashb, fit=(xl) (Top) (vxl1) (Bottom) (backwardlabel) (vxl)] {};

\end{tikzpicture}}
    }
    \caption{The three operations associated with training a layer with weights that perform the function $\mathcal{F}$: Forward activation calculation, backward gradient propagation, and the weight update.
    }
    \label{fig:training-step}
\end{figure}

If we apply sparsity to a linear layer (i.e., $\mathcal{F}$ is a fully-connected layer), our aim is to control:
\begin{enumerate}
    \item \textbf{Forward pass:} $\YB = \mathcal{F}(\XB, \WB \odot \MB) = \XB(\WB \odot \MB)$.
    \item \textbf{Backward pass:} $\nabla_{\XB} \mathcal{L} = (\nabla_{\YB} \mathcal{L}) \cdot (\WB \odot \MB)^\top$.
    \item \textbf{Effect of weight update $\Delta \WB$ on $\YB$:} $\Delta \YB = \XB (\Delta \WB \odot \MB)$\footnote{After a weight update $\Delta \WB$ is applied, new output activations can be written as $\YB + \Delta \YB = \XB (\WB \odot \MB) + \XB (\Delta \WB \odot \MB)$. Our goal is to control $\Delta \YB$.}.
\end{enumerate}

\subsection{Feature learning: Defining the goal of \ac{mup} and \ac{smup}} \label{sec:fld}
Prior works \cite{yang2020feature, yang2022mup, yang2023spectral} introduce the Feature Learning Desiderata (FLD) to ensure stable training dynamics as width is varied. Building on prior works, we include gradients $\nabla_{\XB} \mathcal{L}$ in the desiderata. 

\desiderata{\textbf{Feature Learning Desiderata (FLD):} For layer $l$ and token $i$, we desire that $\| \YB^l_i \|_2 = \Theta(\sqrt{d_\text{out}}), \| \nabla_{\XB} \mathcal{L}^l_i \|_2 = \Theta(\sqrt{d_\text{in}}), \| \Delta \YB^l_i \|_2 = \Theta(\sqrt{d_\text{out}}), \forall i, \forall l$.}

Recall that if all the entries of some vector $\mathbf{v} \in \mathbb{R}^n$ are some constant $c$, then $\| \mathbf{v} \|_2=\Theta(\sqrt{n})$ with respect to width $n$.
Therefore we can satisfy the FLD by ensuring the \emph{typical element size} of $\YB$, $\nabla_{\XB} \mathcal{L}$, and $\Delta \YB$ is $\Theta(1)$ with respect to \textbf{some variable(s)} we would like to scale. Variables to scale include width \cite{yang2020feature, yang2022mup, yang2023spectral}, depth \cite{yang2023feature, bordelon2024depthwise}, and sparsity (this work). The FLD prescribes a \textbf{holistic} signal propagation approach of controlling each of the three operations in a training step, not a subset\footnote{For example, initialization methods alone can only control $\| \YB \|_F$ and $\| \nabla_{\XB} \mathcal{L} \|_F$ at the first time step.}.

\subsection{Maximal update parameterization (\textmu P)} \label{sec:mup}
Here we provide a brief overview of \acf{mup} \cite{yang2020feature, yang2022mup, yang2023spectral}.
With the standard parameterization (SP), \citet{yang2020feature} show the scale of activations throughout training increases as model width increases, motivating the development of \ac{mup}.
\Ac{mup} \cite{yang2020feature, yang2022mup} is defined as the unique parameterization that satisfies the FLD by ensuring the \emph{typical element size} of $\YB$, $\nabla_{\XB} \mathcal{L}$, and $\Delta \YB$ is $\Theta(1)$ \textbf{with respect to change in width $m_d$}. The FLD can also be satisfied by controlling the spectral norm of weights \cite{yang2023spectral}. \ac{mup} enables \textmu Transfer: the optimum learning rate, initialization weight variance, scalar multipliers, and learning rate schedule all remain consistent as width is increased for \textmu P models \cite{yang2022mup}. \textmu Transfer can be leveraged to take a \emph{tune small, train large} approach where hyperparameters are extensively tuned for a small model then transferred, enabling reduced tuning budgets and superior tuning for large models compared to standard practice.

\subsection{Sparse maximal update parameterization (\ac{smup})} \label{sec:smup-overview}

\begin{figure}[h]
    \centering
    \begin{tikzpicture}[
        title/.style={font=\sffamily\bfseries}
    ]
        \node[anchor=south west,inner sep=0] (image) at (0,0) {
            \includegraphics[width=0.9\linewidth]{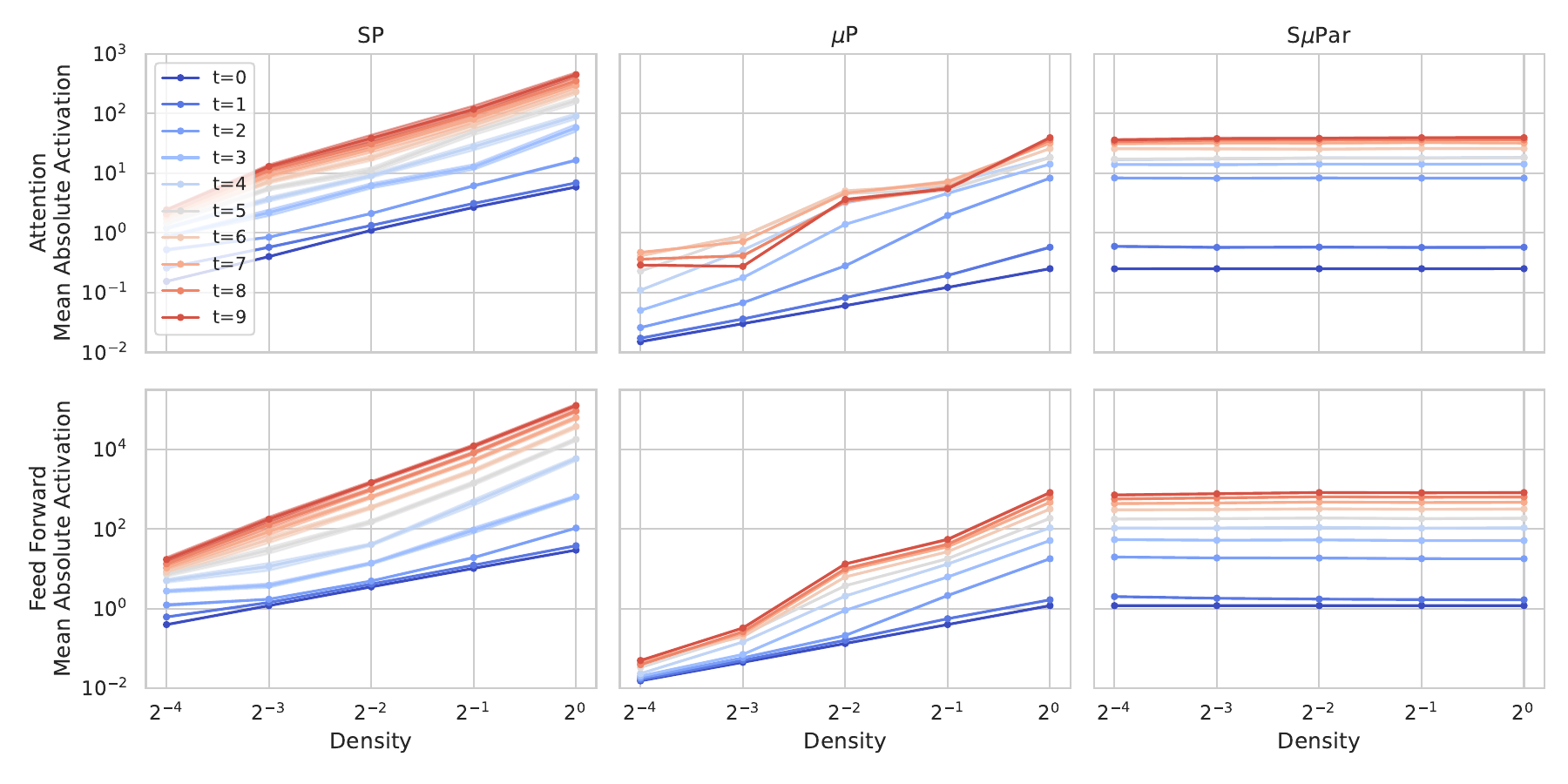}
        };
        \begin{scope}[x={(image.south east)},y={(image.north west)}]
            \draw[-latex, thick, black] (0.58,0.875) -- (0.48,0.875) node[midway, above] (arr1) {};
            \node[title, below=0.05cm of arr1] {\tiny increasing sparsity};
        \end{scope}
    \end{tikzpicture}
    \caption{Mean absolute value activations for attention and feed forward blocks after training step $t$ (10 seeds). In SP and \ac{mup} models, decreasing density causes activations to vanish (note axes on log-scale). In \ac{smup} models, density has little effect on activation scales and there is no vanishing.}
    \label{fig:coord-check}
\end{figure}

\citet{yang2022mup} show activation magnitudes explode with increasing model width. In Figure \ref{fig:coord-check} we show sparsity has the opposite effect: increasing sparsity causes shrinking activation magnitudes.

\result{Increasing sparsity causes vanishing activations and gradients with both SP and \ac{mup}.}
\Ac{smup} is defined as the unique parameterization that satisfies the FLD by ensuring the \emph{typical element size} of $\YB$, $\nabla_{\XB} \mathcal{L}$, and $\Delta \YB$ is $\Theta(1)$ \textbf{with respect to change in width $m_d$ and change in density $m_\rho$}. \Ac{smup} enables stable activation scales across sparsity levels (Figure \ref{fig:coord-check}, right). In this section, we walk through the changes required to control each of the three operations in a sparse training step, providing an overview of the \ac{smup} derivation. We focus on the AdamW \cite{AdamW} optimizer used in our experiments.  For a more detailed derivation, including both SGD and Adam, see Appendix~\ref{sec:smup-detailed-derivation}.

\paragraph{Forward pass at initialization}
To ensure the \emph{typical element size} of $\YB$ is $\Theta(1)$ with respect to change in width $m_{d_\text{in}}$ and change in density $m_\rho$, we can control the mean and variance of $\YB_{ij}$. Since at initialization $\E[\WB]=0$, $\E[\YB] = 0$, and $\WB \perp \YB$, the mean is controlled. The variance of $\YB_{ij}$ can be written as:
\begin{align}
    \Var(\YB_{ij}) = m_{d_\text{in}} d_{\text{in,base}} m_{\rho} \rho_{\text{base}} \sigma^2_{W} (\Var(\XB) + \E[\XB]^2)
\end{align}
To ensure $\Var(\YB_{ij})$ scales independent of $m_{d_\text{in}}$ and $m_{\rho}$, we choose $\sigma^2_{\WB} = \frac{\sigma_{\WB,base}^2}{m_{d_\text{in}} m_{\rho}}$.

\paragraph{Backward gradient pass at initialization}
To ensure the \emph{typical element size} of $\nabla_{\XB} \mathcal{L}$ is $\Theta(1)$ with respect to change in width $m_{d_\text{out}}$ and change in density $m_\rho$, we can control the mean and variance of $\nabla_{\XB} \mathcal{L}$. Since at initialization $\E[\WB]=0$, $\E[\nabla_{\XB} \mathcal{L}] = 0$ and the mean is controlled\footnote{Although the gradients $\nabla_{\YB} \mathcal{L}$ will have some correlation with weights $\WB$ even at initialization, we assume for simplicity that they are fully independent. Future work could investigate this assumption more deeply.}. The variance of $\nabla_{\XB} \mathcal{L}_{ij}$ can be written as:
\begin{align}
    \Var(\nabla_{\XB} \mathcal{L}_{ij}) = m_{d_\text{out}} d_{\text{out,base}} m_{\rho} \rho_{\text{base}} \sigma^2_{\WB}\Var(\nabla_{\YB} \mathcal{L})
\end{align}
To ensure $\Var(\nabla_{\XB} \mathcal{L}_{ij})$ scales independent of $m_{d_\text{out}}$ and $m_{\rho}$, we choose $\sigma^2_{\WB} = \frac{\sigma_{\WB,base}^2}{m_{d_\text{out}} m_{\rho}}$. Typically $m_{d_\text{out}} = m_{d_\text{in}}$, allowing the same $\sigma^2_{\WB}$ to control both forward and backward scales.

\paragraph{Effect of Adam weight update $\Delta \WB$ on $\YB$}
To ensure the \emph{typical element size} of $\Delta \YB$ is $\Theta(1)$ with respect to change in width $m_{d_\text{out}}$ and change in density $m_\rho$. By the law of large numbers, the expected size of each element can be written as:
\begin{align}
    \E[\Delta \YB_{ij}] \to \eta m_{d_\text{in}}d_{\text{in,base}}m_{\rho} \rho_{\text{base}} \E \left[ \XB_{ik} \left( \frac{\sum^T_t \gamma_t \sum_b^B \XB^{t}_{bk} \nabla_\YB \mathcal{L}^{t}_{bj} }{\sqrt{\sum_t^T \omega_t \sum_b^B (\XB^{t}_{bk} \nabla_\YB \mathcal{L}^{t}_{bj})^2}} \right) \right], \text{ as } (d_\text{in}\rho) \to \infty
\end{align}
To ensure $\Delta \YB_{ij}$ and $\|\Delta \YB\|_F$ are scale invariant to $m_{d_\text{in}},m_{\rho}$, we choose $\eta = \frac{\eta_{\text{base}}}{m_{d_\text{in}}m_{\rho}}$.

\paragraph{Implementation summary} \label{sec:implementation-summary}
Table \ref{table:implementation-summary} summarizes the differences between SP, \ac{mup}, and \ac{smup}. Since we only sparsify hidden weights, \ac{smup} matches \ac{mup} for input, output, bias, layer-norm, and attention logits. Also note width multipliers $m_d$ and density multipliers $m_\rho$ are usually the same for all layers, allowing simplified notation. This correction is equivalent to \textmu P \cite{yang2022mup} when $\rho=1$ and $m_{\rho}=1$. The correction to hidden weight initialization we derive is similar to the sparsity-aware initialization in prior work \cite{liu2018rethinking,ramanujan2020what,evci2022gradient}. \Ac{smup} should also easily extend to 2:4 sparsity patterns because, in expectation, the rows and columns of $M^l$ should have equal density. A minimal implementation of \ac{smup} is available at \url{https://github.com/EleutherAI/nanoGPT-mup/tree/supar}.

\begin{table}[h]
    \caption{Summary of SP, \textmu P, and \ac{smup} implementations.}
    \label{table:implementation-summary}
    \centering
    \begin{tabular}{lccc}
        \toprule
        Parameterization    & SP & \begingroup\color{orange}\ac{mup}\endgroup & \begingroup\color{ForestGreen}\ac{smup}\endgroup \\
        \midrule
        Embedding Init. Var.      & $\sigma_{\text{base}}^2$ 
                            & $\sigma_{\text{base}}^2$ 
                            & $\sigma_{\text{base}}^2$ \\
        Embedding LR                & $\eta_{\text{base}}$
                                    & $\eta_{\text{base}}$
                                    & $\eta_{\text{base}}$\\
        Embedding Fwd. & $\XB^{0} \WB_{\text{emb}}$
                                & $\begingroup\color{orange}\alpha_{\text{input}}\endgroup \cdot \XB^{0} \WB_{\text{emb}}$
                                & $\begingroup\color{orange}\alpha_{\text{input}}\endgroup \cdot \XB^{0} \WB_{\text{emb}}$ \\
        Hidden Init. Var. & $\sigma_{\text{base}}^2$
                                 & $\sigma_{\text{base}}^2/ \begingroup\color{orange}m_d\endgroup$
                                 & $\sigma_{\text{base}}^2/ (\begingroup\color{orange}m_d\endgroup \begingroup\color{ForestGreen}m_\rho\endgroup)$ \\
        Hidden LR (Adam)         & $\eta_{\text{base}}$
                                & $\eta_{\text{base}} / \begingroup\color{orange}m_d\endgroup$
                                & $\eta_{\text{base}} / (\begingroup\color{orange}m_d\endgroup \begingroup\color{ForestGreen}m_\rho\endgroup)$ \\
        Unembedding Fwd. & $\XB^{L} \WB^\top_{\text{emb}}$
                                 & $\begingroup\color{orange}\alpha_{\text{output}}\endgroup \XB^{L} \WB^\top_{\text{emb}} / \begingroup\color{orange}m_d\endgroup$
                                 & $\begingroup\color{orange}\alpha_{\text{output}}\endgroup \XB^{L} \WB^\top_{\text{emb}} / \begingroup\color{orange}m_d\endgroup$ \\
        Attention logits            & $\QB^\top \KB / \sqrt{d_{\text{head}}}$
                                & $\QB^\top \KB / \begingroup\color{orange}d_{\text{head}}\endgroup$
                                & $\QB^\top \KB / \begingroup\color{orange}d_{\text{head}}\endgroup$ \\
        \bottomrule
    \end{tabular}
\end{table}

\section{\ac{smup} Training Results}

Here, we present empirical results showing the effectiveness of \ac{smup} over SP and \ac{mup} when training sparse models. When using SP or \ac{mup}, optimal HPs drift as we change the sparsity level, possibly leading to inconclusive or even reversed findings. \ac{smup} has stable optimal HPs across both model width and sparsity level, and we show it improves over SP and \ac{mup} across different scaling approaches. Taken together, we see that \ac{smup} sets the Pareto frontier best loss across all sparsities and widths, including when we scale to a large dense model with width equal to GPT-3 XL~\cite{brown2020language}. Optimal \emph{dense} \ac{mup} HPs---when adjusted using \ac{smup}---are also optimal HPs for all sparse models that we test here.

All tests in this section use GPT-like transformer language models~\cite{radford2019gpt2, dey2023btlm3b8k}, trained on the SlimPajama dataset~\cite{cerebras2023slimpajama} with a 2048 token context length. We apply random unstructured static sparsity to all projection weights in attention and feedforward blocks while keeping embedding, layer normalization, and bias parameters dense. We refer the reader to Appendix~\ref{sec:experiment-details} for full methodology of all experiments.

\subsection{Sparse hyperparameter transfer}
\begin{figure}[h]
    \centering
    \includegraphics[width=0.9\linewidth]{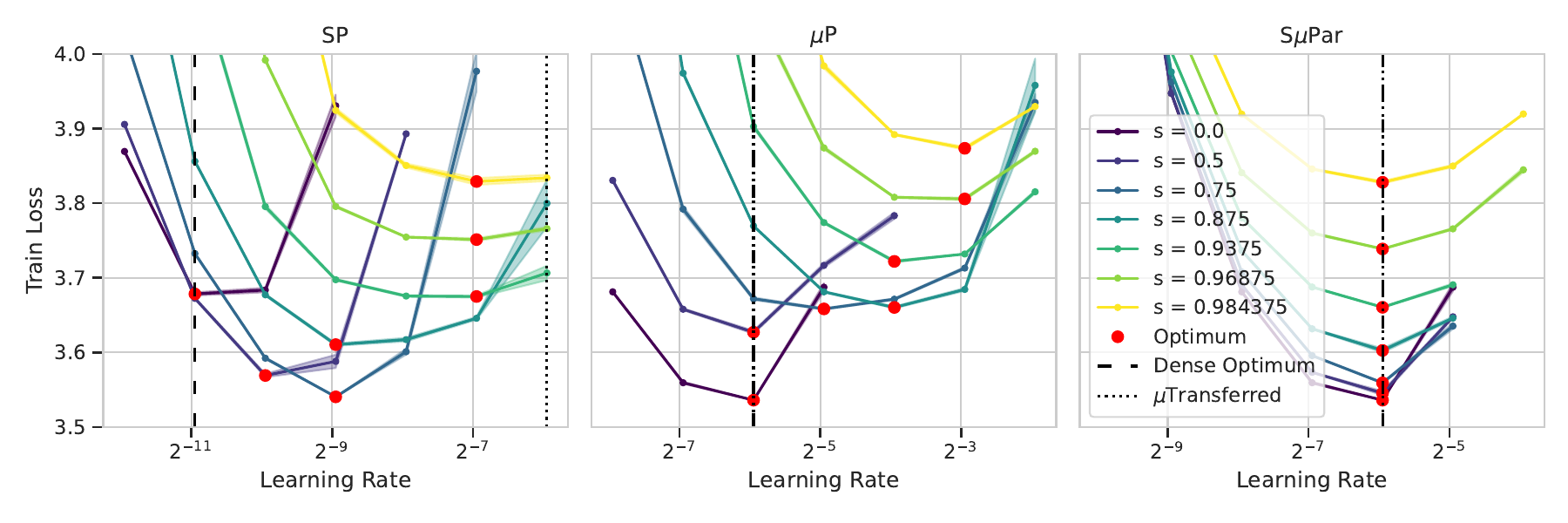}
    \caption{\ac{smup} ensures stable optimal learning rate for any sparsity $s$, unlike SP and \ac{mup} (3 seeds).}
    \label{fig:mutransfer-lr}
\end{figure}

We first show sparsifying a dense model using either SP or \ac{mup} leads to significant drift in optimal HPs as the sparsity level changes. Figure~\ref{fig:mutransfer-lr} shows train loss for SP, \ac{mup}, and \ac{smup} models when trained with varying sparsity levels and sweeping across different peak learning rates.
For the SP configuration, as sparsity increases, the
optimal learning rate increases in a somewhat unpredictable way.
\ac{mup} experiences similar shift in optimal learning rate, though shifts are even more abrupt.
For \ac{smup}, the optimal learning rate is consistently near $2^{-6}$ across all sparsity levels.

\begin{figure}[h]
    \centering
    \includegraphics[width=0.9\linewidth]{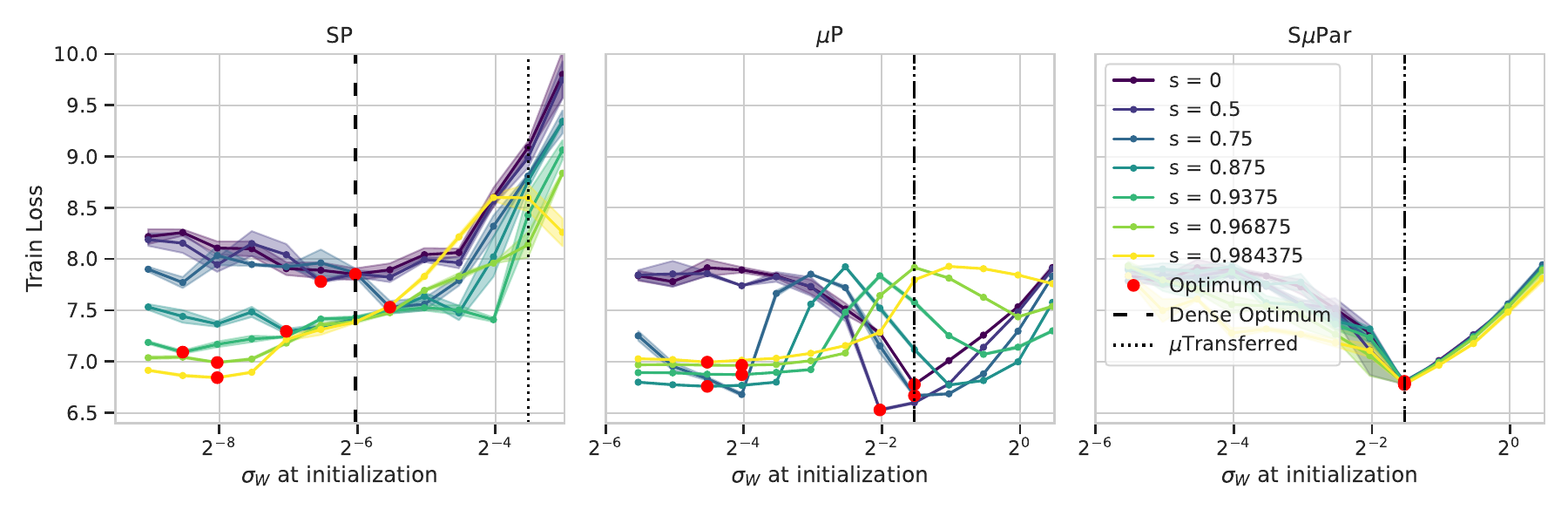}
    \caption{Across sparsity $s$, SP and \ac{mup} show unstable optimal initialization. \ac{smup} is stable (3 seeds).}
    \label{fig:mutransfer-init}
\end{figure}

We also sweep base weight initialization values and find even more chaotic behaviors for SP and \ac{mup} with different sparsity levels (Figure~\ref{fig:mutransfer-init}, left and center, respectively)\footnote{These results are taken from a point early in training as models with widely varying initialization tend to become unstable later in training.}. \ac{mup} even shows discontinuities in optimal initialization values at different sparsity levels. We attribute this discontinuity to widely varying expected activation scales between embedding and transformer decoder layers, where embedding activation scales will tend to dominate for high sparsity levels. \ac{smup} shows consistent optimal initialization (right plot). Figures~\ref{fig:mutransfer-lr} and \ref{fig:mutransfer-init} demonstrate our second finding.

\result{With SP and \ac{mup}, dense and sparse networks do not share the same optimal HPs.}

\begin{wrapfigure}{r}{0.4\textwidth}
    \vspace{-0.6cm}
    \centering
    \includegraphics[width=0.95\linewidth]{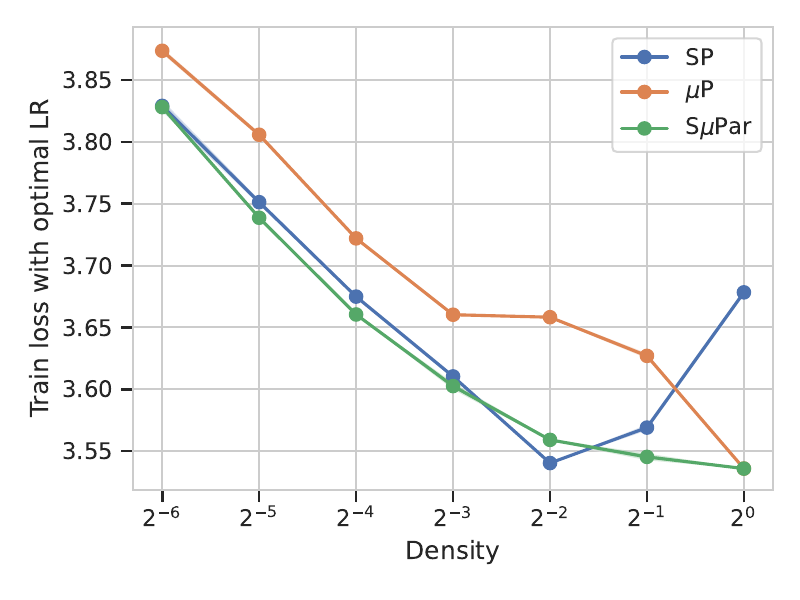}
    \vspace{-0.2cm}
    \caption{Summarizing loss results from Figure~\ref{fig:mutransfer-lr} with the optimal learning rate for each parameterization and sparsity.}
    \label{fig:smup_pareto_frontier}
    \vspace{-10pt}
\end{wrapfigure}

Figure~\ref{fig:smup_pareto_frontier} summarizes our HP transfer tests, showing loss for each parameterization across all sparsities. Even when selecting the best learning rate at each sparsity level for SP and \ac{mup}, \ac{smup} (largely) forms the Pareto frontier with an average gap of 0.8\% better than SP and 2.1\% better than \ac{mup}.

\subsection{Studying \ac{smup} indicates how some sparse scaling techniques appear to work}
So far, we see \ac{smup} can transfer optimal HPs across sparsity levels, but we have also designed it to transfer HPs across different model widths (hidden sizes), similar to \ac{mup}. Here, we further demonstrate that \ac{smup} transfers optimal HPs across width. More generally, sparse scaling that keeps a fixed number of non-zero weights per neuron allows SP and \ac{mup} to also transfer HPs.

\begin{figure}[t]
    \centering
    \includegraphics[width=0.9\linewidth]{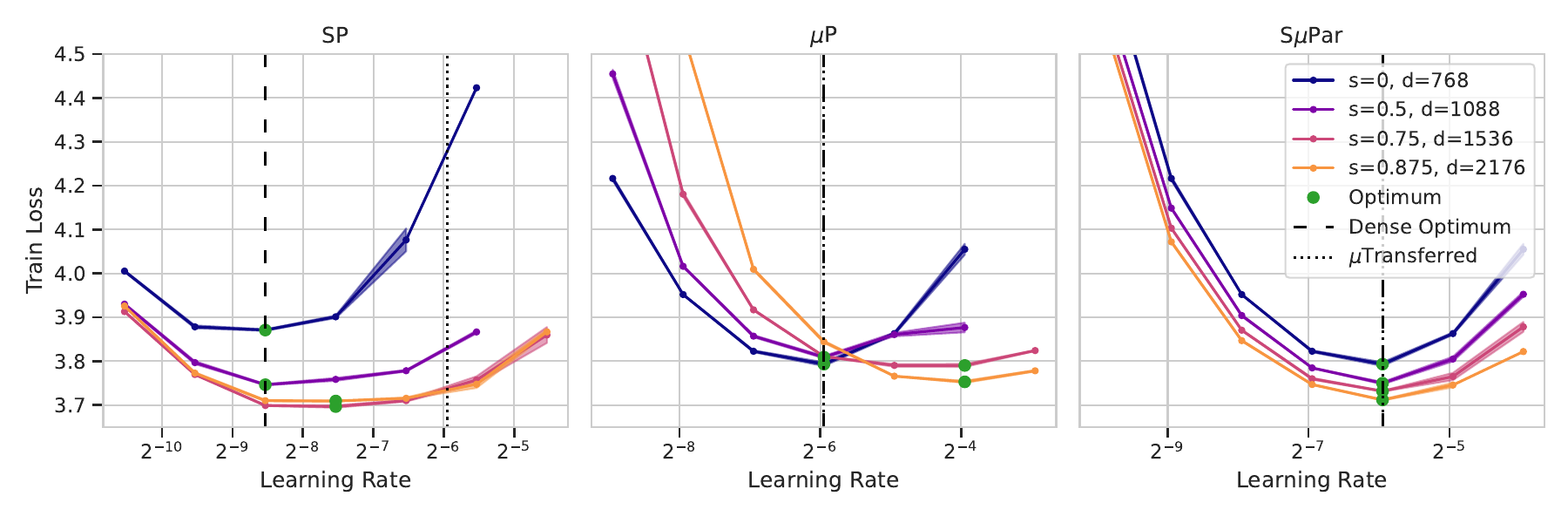}
    \caption{\ac{smup} ensures stable optimal learning rate in Iso-Parameter sparse + wide scaling (3 seeds).}
    \label{fig:mutransfer-lr-iso-flop}
\end{figure}

Figure~\ref{fig:mutransfer-lr-iso-flop} shows learning rate transfer tests when changing both the model's hidden size, $d_{\text{model}}$, and sparsity level in a common scaling approach called \emph{Iso-Parameter scaling} \cite{golubeva2020wider, evci2020rigging, thangarasa2023sparse}. Iso-Parameter scaling keeps the model's number of non-zero parameters approximately the same, as width and sparsity are varied\footnote{Not perfectly Iso-Parameter due to unsparsified layers (embedding, bias, layer-norm, etc.)}. Here, we see the common result that SP models starting from dense HPs \emph{do} tend to significantly improve as we increase width and sparsity. Note, though, the optimal learning rate for each sparsity level still shifts. When we correct dense HPs using \ac{mup} or \ac{smup}, the dense baseline significantly improves, but only \ac{smup} shows consistent loss improvement and stable HPs.

\begin{figure}[h]
\centering
\begin{minipage}{.485\textwidth}
    \centering
    \includegraphics[width=0.8\linewidth]{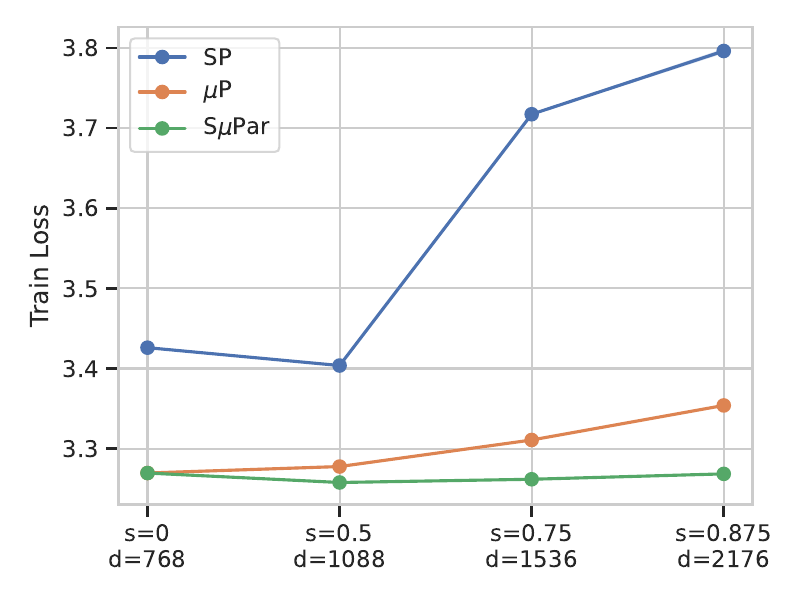}
    \captionof{figure}{Losses at the end of training when Iso-Parameter scaling.}
    \label{fig:demo-iso-flop}
\end{minipage}\hfill
\begin{minipage}{.485\textwidth}
    \includegraphics[width=0.8\linewidth]{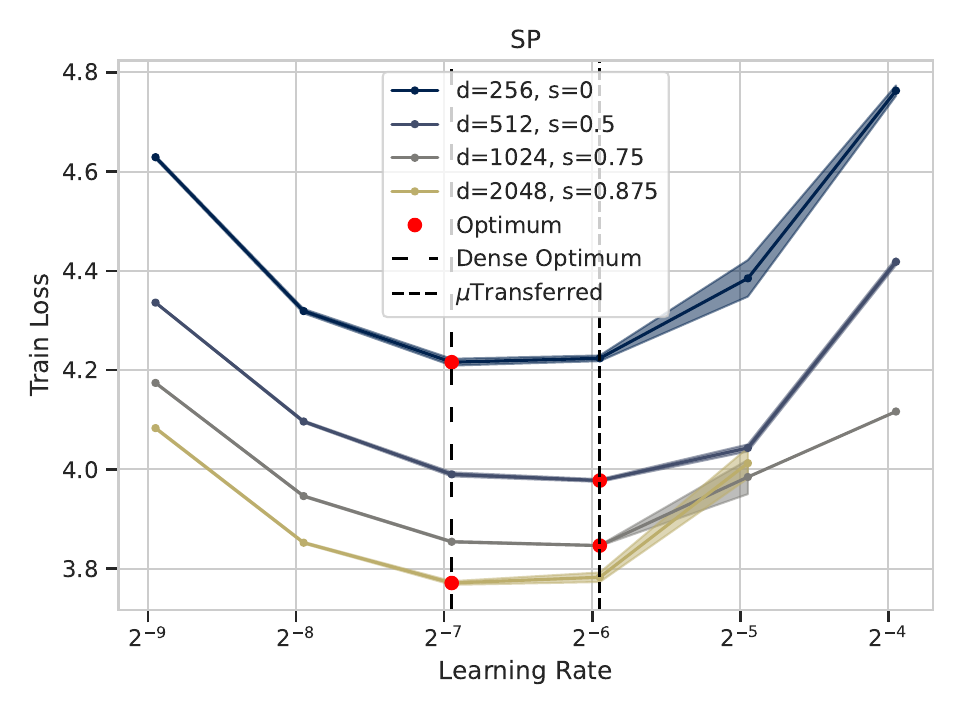}
    \caption{The SP optimized LR is relatively stable with iso-WPN scaling (3 seeds).}
    \label{fig:mutransfer-lr-iso-wpn}
\end{minipage}
\end{figure}

Based on the \ac{smup} formulation: When the number of non-zero weights per neuron (WPN) in the network is the same, \ac{mup} and \ac{smup} become synonymous, because initialization and learning rate adjustment factors will be constant (i.e., $d_{\text{model}}\cdot\rho = \text{WPN} = O(1)$). Optimized SP HPs will also tend to work well. We define this new scaling setting, which we call Iso-WPN, to verify this hypothesis. In Figure~\ref{fig:mutransfer-lr-iso-wpn}, we test SP HPs with Iso-WPN scaling and see the optimal learning rate stays consistently between $2^{-7}$ and $2^{-6}$ with roughly aligned curves (we omit similar \ac{mup} and \ac{smup} plots for space, because their corrections are the same). The conclusion is that when scaling SP models in an Iso-WPN sparse setting, HPs should maintain similar training dynamics. More generally, as WPN decreases (e.g., by increasing sparsity), the optimal learning rate will tend to increase proportionally, and vice versa\footnote{Our results generalize the Yang et al. finding that optimal LR decreases as width increases~\cite[Figure 1]{yang2022mup}.}.

Figures~\ref{fig:coord-check}, \ref{fig:mutransfer-lr}, \ref{fig:mutransfer-init}, and \ref{fig:mutransfer-lr-iso-flop} show \ac{smup} is the only parameterization that ensures stable activation scales and stable optimal HPs across model widths and sparsities, satisfying the FLD.
\result{\ac{smup} enables stable activation and stable optimal HPs for any width and sparsity.}

\subsection{\ac{smup} scaling to large language model pretraining} \label{sec:demo}

We conclude this section reflecting on the demonstration of \ac{smup} improvements in a large-scale language model. We train 610M parameter models starting from a Chinchilla~\cite{hoffmann2022chinchilla} compute-optimal training configuration with 20 tokens per parameter from the SlimPajama dataset. This larger model---with hidden size 2048, 10 layers, and attention head size 64---permits sweeping over a larger range of sparsity levels, so we test up to 99.2\% sparsity (density $2^{-7}$).

Figure~\ref{fig:demo-reduced-dense-20tpp} shows validation loss for each parameterization as we sweep sparsity levels. Additionally, in Table \ref{tab:downstream}, we evaluate the models from Figure \ref{fig:demo-reduced-dense-20tpp} on five downstream tasks: ARC-easy, lambada, RACE, PIQA, and BoolQ, which collectively test for common sense reasoning, world knowledge, and reading comprehension. As sparsity increases, results across pretraining loss and average downstream task accuracy consistently show SP and \ac{mup} fall farther behind \ac{smup}. Since these models are trained with a large number of tokens, we attribute the widening loss gap mostly to increasingly under-tuned learning rates for SP and \ac{mup} as sparsity increases--the weight updates lose gradient information throughout training. Figure \ref{fig:smup_pareto_frontier} shows retuning SP and \ac{mup} could recover some of the gap to \ac{smup}, but that could be costly: These runs take 3-6 hours on a Cerebras CS-3 system (or $>9$ days on an NVIDIA A100 GPU).

Finally, returning to the Iso-Parameter scaling setting, Figure~\ref{fig:demo-iso-flop} shows losses for 111M parameter models trained on 1B tokens and scaled up while using dense optimal HPs. The SP and \ac{mup} models experience detuning as sparsity increases, allowing \ac{smup} to achieve superior losses\footnote{Note this is not an Iso-FLOP comparison because increasing $d_\text{model}$ also increases attention dot product and embedding FLOPs, which aren't be sparsified. This is so significant that our 87.5\% sparse model from Figure~\ref{fig:demo-iso-flop} has double the training FLOPs of the dense baseline, with virtually unchanged loss.}.

\result{Sparse networks trained with \ac{smup} improve over SP and \ac{mup} due to improved tuning.} 

\subsection{Dynamic sparsity hyperparameter transfer} \label{sec:dst}
\begin{figure}[h]
    \centering
    \includegraphics[width=0.9\linewidth]{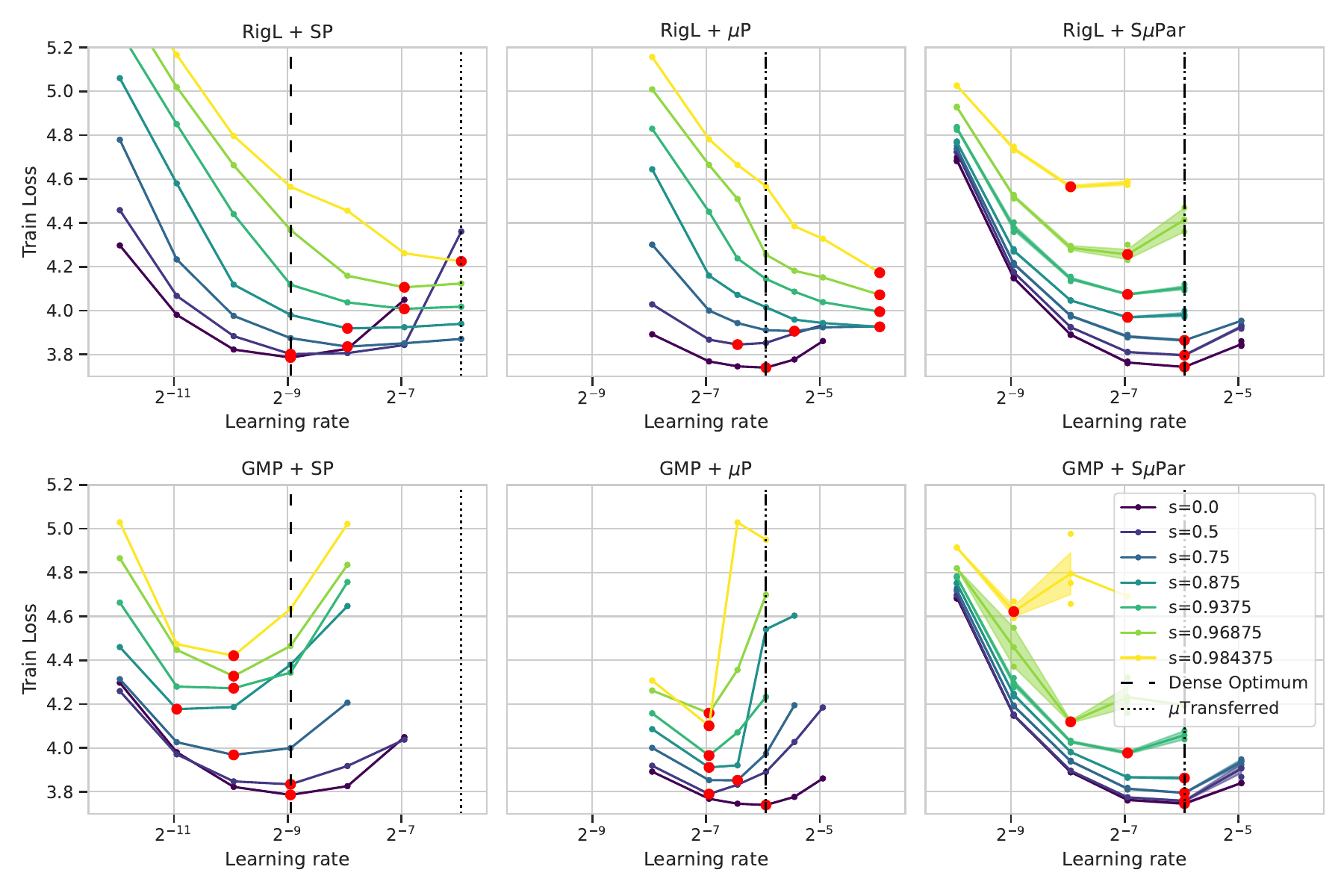}
    \caption{For dynamic sparse training methods RigL and GMP, none of SP, \ac{mup}, or \ac{smup} achieve stable optimal learning rate across sparsity (3 seeds). Missing points indicate diverged training runs.}
    \label{fig:dst-mutransfer-lr}
\end{figure}

In Figure \ref{fig:dst-mutransfer-lr} we test the transfer of optimal learning rate across sparsity levels for two popular dynamic sparse training methods: Rigging the Lottery (RigL) \cite{evci2020rigging}\footnote{RigL: Uniform sparsity distribution, drop fraction of 0.3, and mask updates every 100 steps.} and Gradual Magnitude Pruning (GMP) \cite{zhu2017prune}\footnote{GMP: Uniform sparsity distribution, cubic sparsity schedule, and mask updates every 100 steps.}. We show that none of SP, \ac{mup}, or \ac{smup} achieve transfer of optimal learning rate across sparsity levels. For SP and \ac{mup} we see that higher sparsity levels have higher optimal learning rates. This is because sparsity reduces activation and gradient scales such that a larger learning rate is needed to counteract this. \ac{smup} sees the opposite trend where higher sparsity levels have lower optimal learning rates, indicating that \ac{smup} is ``overcorrecting''.

Dynamic sparse methods can make updates to the weight mask such that the distribution of unmasked/non-zero weights changes to something non-Gaussian, which prevents \ac{smup} from being correct in expectation. Compared to random pruning, a mask obtained from magnitude pruning will better preserve the size of activations and gradients seen in the dense network. Since \ac{smup} assumes weights are drawn from a Gaussian distribution, \ac{smup} ends up ``overcorrecting'' the initialization and learning rate. In future work it would be impactful to develop a parameterization which generalizes \ac{smup} to work for an arbitrary sparse training algorithm.

\section{Discussion}\label{sec:discussion}
To improve sparse training, prior works make targeted corrections which arise from observations that sparsity can cause degraded activation, gradient, and/or weight update signal propagation. We review these observations and corrections to advocate for holistic control of sparse training dynamics.

\paragraph{Sparsifying Can Cause Vanishing Activations}

\citet{evci2022gradient} note that by initializing weights using dense methods (e.g.,~\cite{glorot2010initialization,he2015delving}), the ``vast majority'' of sparse networks have vanishing activations. \citet[App.~A]{lasby2023dynamic} analyze activation variance as a guide for selecting structured sparsity. The FLD suggest activation norms be measured and controlled with respect to sparsity, so activation variance can be considered a proxy to whether sparsity might negatively impact training dynamics. \citet{evci2022gradient} ultimately initialize variances via neuron-specific sparse connectivity, while \citet{liu2018rethinking} and \citet{ramanujan2020what} propose scaling weight variances proportional to layer sparsity. These corrections, however, only target controlling activations but not weight updates.

\paragraph{Gradient Flow Partially Measures the Weight Update \textmu Desideratum}
Sparsity also impairs \emph{gradient flow}---the magnitude of the gradient to the weights---during training~\cite{evci2022gradient,bambhaniya2024progressive}. Since gradient flow is measured using the norm of the weight gradients, it measures a piece of the weight update. Unfortunately, gradient flow does not directly measure the effect of the weight update step, which can also involve adjustments for things like optimizer state (e.g., momentum and velocity), the learning rate, and weight decay.
Prior works propose techniques to improve gradient flow during sparse training and pruning by adjusting individual hyperparameters or adding normalization~\cite{wang2020picking,lubana2020gradient,evci2022gradient,bambhaniya2024progressive}. However, these techniques might overlook the effects of the optimizer and learning rates in weight updates. Notably, \citet{tessera2021keep} \emph{do} consider some of these effects, but their proposed techniques maintain gradient flow only in the Iso-Parameter scaling setting rather than arbitrary sparsification.

\citet[App.~A.1]{frantar2023scaling} also endeavor to control weight updates, where they observe diminished step sizes when optimizing sparse networks with Adafactor~\cite{shazeer2018adafactor}. They correct this by computing Adafactor's root-mean-square scaling adjustments over \emph{unpruned} weights and updates. However, such normalization does not prevent activations from scaling with model width~\cite{yang2022mup,yang2023spectral}. In this sense, sparsity-aware fixes to Adafactor can improve dynamics, but will not address instability holistically. In Figure \ref{fig:mutransfer-ablation-lr} we show the \ac{smup} LR correction alone is not even sufficient to achieve stable optimal $\eta$.

\paragraph{Weight Initialization Only Controls Dynamics at Initialization}
We noted works above that adjust sparse weight initializations~\cite{evci2022gradient,liu2018rethinking,ramanujan2020what}. Additionally, \citet{lee2019signal} explore orthogonal weight initialization~\cite{pennington2017resurrecting}, both before pruning (to ensure SNIP~\cite{lee2018snip} pruning scores are on a similar scale across layers) and after (to improve trainability of the sparse network).
While adjusting weights can improve sparse training dynamics at initialization, such adjustments are insufficient to stabilize signals \emph{after multiple steps of training}, in the same way that standard weight initializations fail to stabilize training of dense networks. In Figure \ref{fig:mutransfer-ablation-init} we show the \ac{smup} init. alone is not even sufficient to achieve stable optimal $\sigma_W$.

\section{Limitations}\label{sec:limitations}

As Section \ref{sec:dst} shows, \ac{smup} requires further extension for dynamic sparse training due to unpredictable changes in weight distributions. The same applies to methods which prune at initialization or after pretraining in a non-random fashion. Iterative magnitude pruning (IMP) is an interesting case since it involves rewinding weights back to their initial values while maintaining the same mask \cite{frankle2018lottery}. If the IMP mask at initialization still allows the non-zero weights to have a Gaussian distribution, then \ac{smup} would apply to this case. Therefore, it's possible \ac{smup} \emph{could} prevent ``HP detuning'' in later IMP iterations, and \emph{potentially} improve IMP losses, though we do not explore this. \ac{smup} would also work for random structured pruning of entire neurons at initialization because this case simply reduces to training with a narrower dense model.

For weight sparsity more generally, the most pressing limitation is
the lack of hardware acceleration~\cite{mishra2021accelerating}.
While new
software~\cite{schultheis2023towards,lasby2023dynamic,magic2024deepsparse}
continues to better leverage existing hardware, the growth of software
and hardware co-design is also
encouraging~\cite{thangarasa2023sparse,cerebras2024train}, as
effective sparsity techniques can be specifically optimized in deep
learning hardware.
But to effectively plan hardware, we need to train and test sparse
prototypes at next-level sizes, at scales where the optimum sparsity level may be higher than in current
networks~\cite{frantar2023scaling}.
Performing such \emph{scaling law}-style studies requires incredible resources
even for dense models with well-established training
recipes~\cite{kaplan2020scalinglaws,hoffmann2022chinchilla}.
As \ac{smup} reduces training and tuning costs, it can help unlock these
studies and guide future hardware design.

Finally, the scaling factors for the weight update need to be derived for each optimizer, which might limit the usability of \ac{smup} in practice. 
For a discussion of the broader impacts of \ac{smup}, see Appendix~\ref{sec:broader-impacts}.

\section{Conclusion}

Nobody said training with sparsity was easy.
We showed that with the standard parameterization and \ac{mup}, increasing sparsity level directly correlates with vanishing activations.
Impaired training dynamics prevent sparse models from sharing the same
optimal hyperparameters, suggesting prior results based on re-use of dense HPs merit re-examination.
In contrast, by holistically controlling the training process,
\ac{smup} prevents vanishing activations and enables HP transfer
(across both width and sparsity).
LLMs trained with \ac{smup} improve over \ac{mup} and the standard
parameterization.
As such, we hope \ac{smup} makes things a little easier for sparsity research going forward.

\section*{Acknowledgements}
We would like to thank Gavia Gray, who provided helpful feedback on the manuscript, and Gurpreet Gosal, who tuned the \textmu Transferred hyperparameters seen throughout the document.

\bibliographystyle{cereb}
\bibliography{main}

\appendix

\section{Broader impacts} \label{sec:broader-impacts}

Sparsity is recognized to reduce carbon emissions~\cite{patterson2021carbon} and offset well-known environmental and financial costs of large model training~\cite{bender2021dangers}. For example, unstructured sparsity can be accelerated by the Cerebras Wafer-Scale Engine\footnote{\url{https://www.cerebras.net/blog/harnessing-the-power-of-sparsity-for-large-gpt-ai-models}} and 2:4 block sparsity can be accelerated by NVIDIA Ampere GPUs\footnote{\url{https://www.nvidia.com/en-us/data-center/ampere-architecture/}}.
There is growing recognition that HP tuning is a key contributor to these costs. HP tuning is costly, possibly undermining equity in AI research due to financial resources~\cite{strubell2019energy}.
During model \emph{re}training, \emph{sensitivity} to HPs also leads to downstream costs~\cite{strubell2019energy}.
\Ac{smup} can reduce these costs and sensitivities and thus improve equity.

Sparsity also has potential drawbacks.
\citet{hooker2019compressed} showed that even when top-line
performance metrics are comparable, pruned networks may perform worse
on specific subsets of the data (including on underrepresented
groups~\cite{hooker2020characterising}), may amplify sensitivity to
adversarial examples, and may be more sensitive to distribution shift.
These sensitivities may depend on the degree of
sparsity~\cite{guo2018sparse}.  It remains an open question whether
such drawbacks occur only with pruning or when training with sparsity
from scratch (as in \ac{smup})~\cite{hoefler2021sparsity}, and how such
sensitivity may impact susceptibility to
misuse~\cite{wei2024jailbroken}.
We require sparsity-specific methods to
detect~\cite{sheng2019woman,nadeem2020stereoset} and
mitigate~\cite{gonen2019lipstick,ouyang2022training} harm.  Moreover,
since many large models are later pruned for deployment,
we recommend testing and documenting in the model
card~\cite{mitchell2019model} any adverse affects of sparsification at
the time of model release.

\section{Downstream task comparison of parameterizations}
In Table \ref{tab:downstream}, we evaluate the models from Figure \ref{fig:demo-reduced-dense-20tpp} on five downstream tasks: ARC-easy, lambada, RACE, PIQA, and BoolQ, which collectively test for common sense reasoning, world knowledge, and reading comprehension. We also specifically chose tasks that are easy enough for even extremely sparse models to significantly outperform random chance.

\begin{table}[h]
\centering
\resizebox{1.0\columnwidth}{!}{%
\begin{tabular}{|l|c|ccc|ccc|ccc|ccc|}
\hline
Sparsity  & -      & \multicolumn{3}{c|}{0}                        & \multicolumn{3}{c|}{0.5}                      & \multicolumn{3}{c|}{0.75}            & \multicolumn{3}{c|}{0.875}            \\
\hline
          & Rand. & SP            & \acs{mup}           & \acs{smup}         & SP            & \acs{mup}           & \acs{smup}         & SP   & \acs{mup}           & \acs{smup}         & SP   & \acs{mup}           & \acs{smup}         \\
\hline
ARC-easy & 0.25   & 0.49          & \textbf{0.51} & \textbf{0.51} & 0.45          & \textbf{0.49} & 0.48          & 0.44 & \textbf{0.45} & \textbf{0.45} & 0.42 & 0.43          & \textbf{0.44} \\
LAMBADA   & 0.00   & 0.32          & \textbf{0.36} & \textbf{0.36} & 0.27          & 0.31          & \textbf{0.32} & 0.22 & 0.27          & \textbf{0.28} & 0.20 & 0.23          & \textbf{0.25} \\
RACE      & 0.25   & \textbf{0.30} & \textbf{0.30} & \textbf{0.30} & 0.29          & \textbf{0.31} & 0.30          & 0.28 & \textbf{0.30} & 0.29          & 0.27 & \textbf{0.28} & \textbf{0.28} \\
PIQA      & 0.50   & \textbf{0.67} & \textbf{0.67} & \textbf{0.67} & 0.63          & 0.65          & \textbf{0.67} & 0.63 & \textbf{0.64} & \textbf{0.64} & 0.62 & \textbf{0.63} & \textbf{0.63} \\
BoolQ     & 0.50   & 0.53          & \textbf{0.57} & \textbf{0.57} & \textbf{0.58} & 0.55          & 0.51          & 0.57 & \textbf{0.62} & 0.52          & 0.61 & \textbf{0.62} & \textbf{0.62} \\ \hline
Avg.      & 0.30   & 0.46          & \textbf{0.48} & \textbf{0.48} & 0.44          & \textbf{0.46} & \textbf{0.46} & 0.43 & \textbf{0.46} & 0.44          & 0.42 & \textbf{0.44} & \textbf{0.44} \\ \hline \hline
Sparsity  & -      & \multicolumn{3}{c|}{0.9375}                   & \multicolumn{3}{c|}{0.96875}                  & \multicolumn{3}{c|}{0.984375}        & \multicolumn{3}{c|}{0.992188}         \\
\hline
          & Rand. & SP            & \acs{mup}           & \acs{smup}         & SP            & \acs{mup}           & \acs{smup}         & SP   & \acs{mup}           & \acs{smup}         & SP   & \acs{mup}           & \acs{smup}         \\
\hline
ARC-easy & 0.25   & 0.41          & 0.40          & \textbf{0.43} & 0.39          & \textbf{0.42} & 0.41          & 0.38 & 0.38          & \textbf{0.41} & 0.37 & \textbf{0.38} & \textbf{0.38} \\
LAMBADA   & 0.00   & 0.19          & 0.20          & \textbf{0.21} & 0.16          & 0.18          & \textbf{0.19} & 0.13 & 0.15          & \textbf{0.17} & 0.12 & 0.13          & \textbf{0.14} \\
RACE      & 0.25   & 0.25          & 0.27          & \textbf{0.28} & 0.24          & 0.25          & \textbf{0.27} & 0.24 & 0.24          & \textbf{0.25} & 0.25 & 0.24          & \textbf{0.26} \\
PIQA      & 0.50   & \textbf{0.61} & \textbf{0.61} & \textbf{0.61} & 0.60          & \textbf{0.61} & 0.60          & 0.59 & \textbf{0.60} & 0.59          & 0.58 & \textbf{0.60} & \textbf{0.60} \\
BoolQ     & 0.50   & \textbf{0.62}          & \textbf{0.62}          & 0.61          & 0.57          & \textbf{0.62} & \textbf{0.62} & 0.45 & \textbf{0.62} & 0.61          & 0.41 & \textbf{0.61} & \textbf{0.61} \\ \hline
Avg.      & 0.30   & 0.42          & 0.42          & \textbf{0.43} & 0.39          & \textbf{0.42} & \textbf{0.42} & 0.36 & 0.40          & \textbf{0.41} & 0.34 & 0.39          & \textbf{0.40} \\ \hline
\end{tabular}
}
    \caption{Downstream evaluation accuracy; higher is better: \ac{smup} performs best or within 0.01 of best across all sparsity levels and tasks, except $boolq$ at 50\% and 75\% sparsity.  Even at 99\% sparsity, \ac{smup} models maintain 40\%$+$ average accuracy, whereas the SP model drops to 34\%, close to the 30\% accuracy of the random baseline.}
    \label{tab:downstream}
\end{table}

\section{Individual ablations of \ac{smup} initialization and learning rate corrections} \label{sec:mutransfer-ablation}
\begin{figure}[h]
    \centering
    \includegraphics[width=0.9\linewidth]{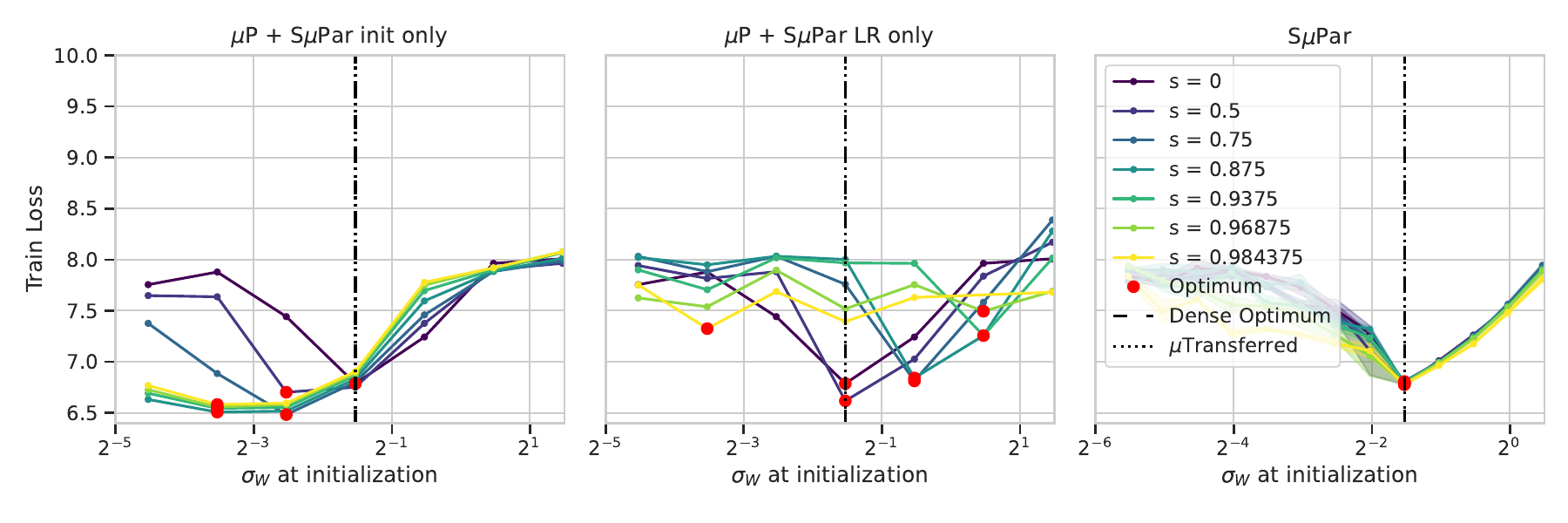}
    \caption{\ac{smup} ensures stable optimal weight initialization standard deviation, unlike SP, \ac{mup}, \ac{mup} + \ac{smup} init only, and \ac{mup} + \ac{smup} LR only.}
    \label{fig:mutransfer-ablation-init}
\end{figure}

\begin{figure}[h]
    \centering
    \includegraphics[width=0.9\linewidth]{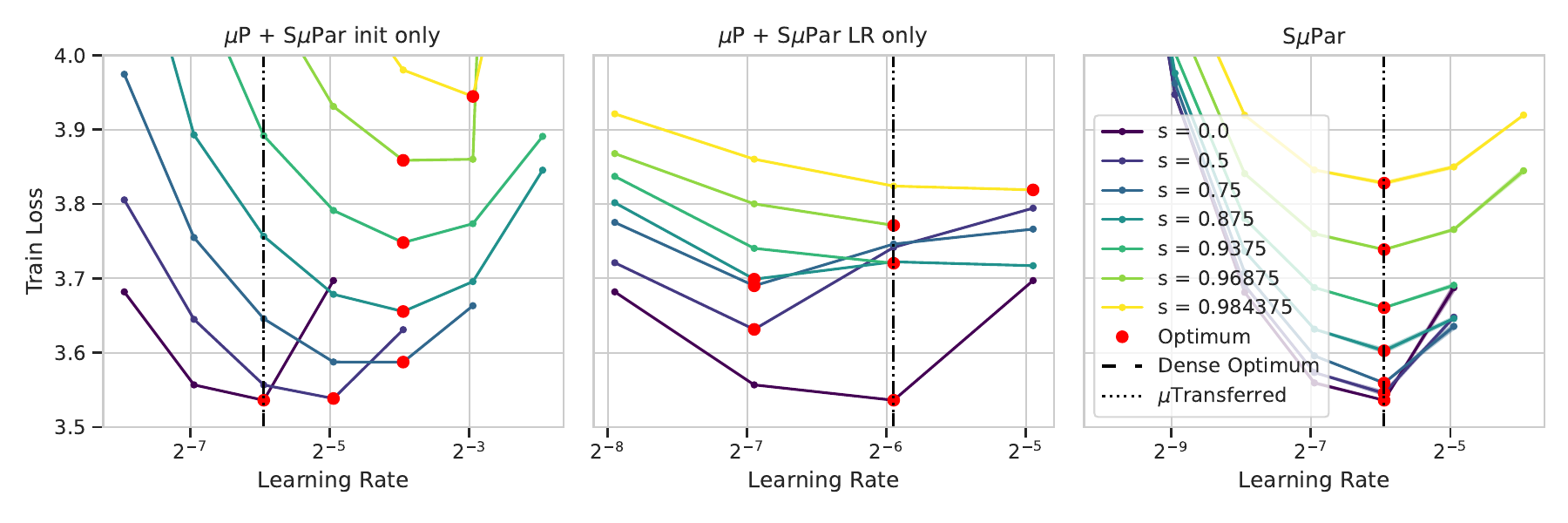}
    \caption{\ac{smup} ensures stable optimal learning rate (\textbf{Bottom}), unlike SP, \ac{mup}, \ac{mup} + \ac{smup} init only, and \ac{mup} + \ac{smup} LR only.}
    \label{fig:mutransfer-ablation-lr}
\end{figure}

In Figures \ref{fig:mutransfer-ablation-init} and \ref{fig:mutransfer-ablation-lr}, we individually ablate the effect of the \ac{smup} initialization and the \ac{smup} learning rate. We show that using only the \ac{smup} initialization in conjunction with \ac{mup} (\ac{mup} + \ac{smup} initialization only) does not allow for transfer of optimal initialization standard deviation or optimal learning rate across sparsity levels. We also show that using only the \ac{smup} learning rate in conjunction with \ac{mup} does not achieve transfer either. Therefore, \textbf{both} the \ac{smup} initialization and learning rate corrections are required to achieve optimal hyperparameter transfer across sparsity levels.


\section{\Ac{smup} detailed derivation} \label{sec:smup-detailed-derivation}

\subsection{Forward pass at initialization}
The first stage where we would like to control training dynamics is in the layer's forward function. For a random unstructured sparsity mask $\MB$, since each \emph{column} of $\MB$ has $d_\text{in}\rho$ non-zero elements in expectation, we can rewrite the forward pass as:
\begin{align}
    \YB_{ij} = \left[ \XB(\WB \odot \MB) \right]_{ij} = \sum_{q=1}^{d_\text{in}} \XB_{iq} (\WB_{qj} \cdot \MB_{qj}) = \sum_{k:\MB_{kj}=1}^{d_\text{in} \rho} \XB_{ik} \WB_{kj}
\end{align} 

To satisfy the FLD, we desire the \emph{typical element size} of $\YB$ is $\Theta(1)$ with respect to change in width $m_{d_\text{in}}$ and change in density $m_\rho$. To achieve this we can ensure the mean and variance of $\YB_{ij}$ are invariant to $m_{d_\text{in}}$ and $m_{\rho}$.

\textbf{Mean:} As expectation is linear and $\XB$ and $\WB$ are independent at initialization:
\begin{align}
    \E[\YB_{ij}] = \E\left[ \sum_{k:\MB_{kj}=1}^{d_\text{in} \rho} \YB_{ik} \WB_{kj} \right] = \sum_{k:\MB_{kj}=1}^{d_\text{in} \rho} \E[\XB_{ik}\WB_{kj}] = \sum_{k:\MB_{kj}=1}^{d_\text{in} \rho} \E[\XB_{ik}] \E[\WB_{kj}]
\end{align}
Therefore, since at initialization $\E[\WB_{ij}]=0$, $\E[\YB_{ij}] = 0$ and the mean is controlled.

\textbf{Variance:} As expectation is linear and each weight element is IID:
\begin{align}
    \Var(\YB_{ij}) = \Var\left(\sum_{k:\MB_{kj}=1}^{d_\text{in} \rho} \XB_{ik} \WB_{kj}\right) = \sum_{k:\MB_{kj}=1}^{d_\text{in} \rho} \Var(\XB_{ik} \WB_{kj})
\end{align}
Then, since $\XB$ and $\WB$ are independent at initialization:
\begin{align}
    \Var(\YB_{ij}) = \sum_{k:\MB_{kj}=1}^{d_\text{in} \rho} (\Var(\XB_{ik}) + \E[\XB_{ik}]^2)(\Var(\WB_{kj}) + \E[\WB_{kj}]^2) - (\E[\XB_{ik}]\E[\WB_{kj}])^2
\end{align}
Finally, since at initialization $\E[\WB_{kj}]=0$ and redefining $\Var(\WB_{kj}) = \sigma^2_{\WB}$:
\begin{align}
    \Var(\YB_{ij}) = \sum_{k:\MB_{kj}=1}^{d_\text{in} \rho} (\Var(\XB_{ik}) + \E[\XB_{ik}]^2)\Var(\WB_{kj}) = d_\text{in}\rho \sigma^2_{\WB} (\Var(\XB) + \E[\XB]^2)
\end{align}
Rewriting in terms of multipliers for the width $m_{d_\text{in}} = \frac{d_\text{in}}{d_\text{in, base}}$ and the change in density $m_\rho = \frac{\rho}{\rho_{\text{base}}}$:
\begin{align}
    \Var(\YB_{ij}) = m_{d_\text{in}} d_\text{in, base} m_\rho \rho_{\text{base}} \sigma^2_{\WB} (\Var(\XB) + \E[\XB]^2)
\end{align}
\textbf{Solution:} To satisfy the FLD and ensure $\Var(\YB_{ij})$ scales independently of $m_{d_\text{in}}$ and $m_\rho$, we choose to set $\sigma^2_{\WB} = \frac{\sigma_{\WB,base}^2}{m_{d_\text{in}}m_\rho}$. This ensures typical entry size of $\YB$ is invariant to changes in width $m_{d_\text{in}}$ and density $m_{\rho}$.

Note that this correction is equivalent to \textmu P \cite{yang2022mup} when $m_\rho=1$. Further, the sparsity factor in the denominator matches the correction for sparsity-aware initialization from \citet{evci2022gradient}.

\subsection{Backward gradient pass at initialization}
The next stage we would like to control training dynamics is in the layer's backward pass. For a random unstructured sparsity mask $\MB$, since each \emph{row} of $\MB$ has $d_\text{out}\rho$ non-zero elements in expectation, we can rewrite the backward pass as:
\begin{align}
    \nabla_{\XB} \mathcal{L}_{ij} = \left[ \nabla_{\YB} \mathcal{L} (\WB \odot \MB)^\top \right]_{ij} =  \sum_q^{d_\text{out}} \nabla_{\YB} \mathcal{L}_{iq} (\WB_{jq} \cdot \MB_{jq}) = \sum_{k:\MB_{jk}=1}^{d_\text{out}\rho} \nabla_{\YB} \mathcal{L}_{ik} \WB_{jk}
\end{align}
To satisfy the FLD, we desire the \emph{typical element size} of $\nabla_{\XB} \mathcal{L}$ is $\Theta(1)$ with respect to change in width $m_{d_\text{out}}$ and change in density $m_\rho$. To achieve this, we can ensure the mean and variance of $\nabla_{\XB} \mathcal{L}$ are invariant to $m_{d_\text{out}}$ and $m_\rho$.

\textbf{Mean:} Although the gradients $\nabla_{\YB} \mathcal{L}$ will have some correlation with weights $\WB$ even at initialization, we assume for simplicity that they are fully independent. Future work could investigate this assumption more deeply. As expectation is linear:
\begin{align}
    \E[\nabla_{\XB} \mathcal{L}_{ij}] = \E\left[\sum_{k:\MB_{jk}=1}^{d_\text{out}\rho} \nabla_{\YB} \mathcal{L}_{ik} \WB_{jk}\right] = \sum_{k:\MB_{jk}=1}^{d_\text{out}\rho} \E[\nabla_{\YB} \mathcal{L}_{ik}\WB_{jk}] = \sum_{k:\MB_{jk}=1}^{d_\text{out}\rho} \E[\nabla_{\YB} \mathcal{L}_{ik}] \E[\WB_{jk}]
\end{align}
Therefore, since at initialization $\E[\WB_{jk}]=0$, $\E[\nabla_{\XB} \mathcal{L}_{ij}] = 0$, the mean is controlled.

\textbf{Variance:} As expectation is linear and each weight element is IID:
\begin{align}
    \Var(\nabla_{\XB} \mathcal{L}_{ij}) = \Var\left(\sum_{k:\MB_{jk}=1}^{d_\text{out}\rho} \nabla_{\YB} \mathcal{L}_{ik} \WB_{jk}\right) = \sum_{k:\MB_{jk}=1}^{d_\text{out}\rho} \Var(\nabla_{\YB} \mathcal{L}_{ik} \WB_{jk})
\end{align}
From the backward pass mean derivation, we know $\E[\nabla_{\YB} \mathcal{L}_{ij}]=0$. Then, similar to the forward pass variance derivation, we can simplify using the facts that at initialization, $\nabla_{\YB} \mathcal{L}$ and $\WB$ are (roughly) independent and $\E[\WB]=0$. Similarly we can also define $\Var(\WB_{kj}^l) = \sigma^2_{\WB}$ and rewrite in terms of width multiplier $m_{d_\text{out}} = \frac{d_\text{out}}{d_{\text{out,base}}}$
and changes in density $m_\rho = \frac{\rho}{\rho_{\text{base}}}$:
\begin{align}
    \Var(\nabla_{\XB} \mathcal{L}_{ij}) = m_{d_\text{out}} d_{\text{out,base}} m_\rho \rho_{\text{base}} \sigma^2_{\WB}\Var(\nabla_{\YB} \mathcal{L})
\end{align}

\textbf{Solution:} To satisfy the FLD and ensure $\Var(\nabla_{\XB} \mathcal{L}_{ij})$ scales independently of $m_{d_\text{out}}$ and $m_\rho$, we choose to set $\sigma^2_{\WB} = \frac{\sigma^2_{\WB,\text{base}}}{m_{d_\text{out}} m_\rho}$. This ensures the typical entry size of $\nabla_{\XB} \mathcal{L}$ is invariant to changes in width $m_{d_\text{out}}$ and density $m_{\rho}$. Typically, we scale model width such that $m_{d_\text{out}} = m_{d_\text{in}}$. This equal scaling allows the same initialization variance to correct both forward activation and backward gradient scales, making them independent of width. Further, since we assume random sparsity across layer's weights, the sparsity initialization correction factor, $m_\rho$, is the same for both the forward activations and backward gradients.

\subsection{Effect of weight update $\Delta \WB$ on $\YB$}

To satisfy the FLD, we desire the \emph{typical element size} of the weight update $\Delta \YB$ is $\Theta(1)$ with respect to change in width $m_{d_\text{in}}$ and change in density $m_\rho$. To achieve this we examine the expected size of each element. Here, we use $\eta$ to be the learning rate for this layer. For a random unstructured sparsity mask $\MB$, since each \emph{column} of $\MB$ has $d_\text{in}\rho$ non-zero elements in expectation:
\begin{align}
    \Delta \YB_{ij} = \left[\eta \XB (\Delta \WB \odot \MB) \right]_{ij} = \eta \sum_{q=1}^{d_\text{in}} \XB_{iq} (\Delta \WB_{qj} \cdot \MB_{qj}) = \eta \sum_{k:\MB_{kj}=1}^{d_\text{in}\rho} \XB_{ik} \Delta \WB_{kj}
\end{align}

\textbf{Mean:} As expectation in linear: 
\begin{align}
    \E[\Delta \YB_{ij}] = \E \left[ \eta \sum_{k:\MB_{kj}=1}^{d_\text{in}\rho} \XB_{ik} \Delta \WB_{kj} \right] = \eta \sum_{k:\MB_{kj}=1}^{d_\text{in}\rho} \E[\XB_{ik} \Delta \WB_{kj}]
\end{align}

Since $\Delta \mathbf{W}$ was derived from $\mathbf{X}$, there is covariance between these variables and $\E[\mathbf{X}_{ik} \Delta \mathbf{W}_{kj}]$ is non-zero.  By the Law of Large Numbers:
\begin{align}
    \E[\Delta \YB_{ij}] \to \eta d_\text{in}\rho \E \left[ \XB_{ik} \Delta \WB \right], \text{ as } (d_\text{in}\rho) \to \infty
\end{align}
Rewriting in terms of width and density multipliers:
\begin{align}
    \label{eqn:weight-update-lln}
    \E[\Delta \YB_{ij}] \to \eta m_{d_\text{in}}d_\text{in,base}m_\rho \rho_\text{base} \E \left[ \XB_{ik} \Delta \WB \right], \text{ as } (d_\text{in}\rho) \to \infty
\end{align}
Equation \ref{eqn:weight-update-lln} will be used as intermediate result in the following sections.

\subsubsection{Effect SGD weight update $\Delta \WB$ on $\YB$}
Following the formulation in \cite{yang2022mup}, stochastic gradient descent (SGD) weight updates take the form: 
\begin{align}
    \Delta \WB^l_{kj} = \left[\frac{(\XB)^\top (\nabla_{\mathbf{Y}} \mathcal{L})}{d_\text{in}} \right]_{kj} = \frac{1}{d_\text{in}} \sum_{b=1}^B \XB_{bk} (\nabla_{\mathbf{Y}} \mathcal{L})_{bj}
\end{align}

so we can rewrite Equation \ref{eqn:weight-update-lln} as:
\begin{align}
    \E[\Delta \YB_{ij}] \to \eta m_\rho \rho_\text{base} \E \left[ \XB_{ik} \sum_{b=1}^B \XB_{bk} (\nabla_{\mathbf{Y}} \mathcal{L})_{bj} \right], \text{ as } (d_\text{in}\rho) \to \infty
\end{align}
\textbf{Solution:} For SGD updates, to satisfy the FLD and ensure $\E[\Delta \YB_{ij}]$ and the typical entry size of $\Delta \mathbf{Y}$ are scale invariant to $m_d$ and $m_\rho$, we choose $\eta = \eta_{\text{base}} / m_\rho$. Note this correction is equivalent to \textmu P \cite{yang2022mup} when $\rho=1,m_\rho=1$.

\subsubsection{Effect of Adam weight update $\Delta \WB$ on $\YB$}
Following the formulation in \citet{yang2022mup}, Adam weight updates take the form: 
\begin{align}
    \Delta \WB_{kj} = \frac{\sum^T_t \gamma_t \sum_b^B \XB^{l,t}_{bk} (\nabla_{\mathbf{Y}} \mathcal{L})^t_{bj} }{\sqrt{\sum_t^T \omega_t \sum_b^B (\XB^t_{bk} (\nabla_{\mathbf{Y}} \mathcal{L})^t_{bj})^2}}
\end{align}
where $T$ is the current training step and $\gamma_t,\omega_t$ are the moving average weights at each training step. Here, we can just consider the weight update associated with an unpruned weight, since a pruned weight will have value and update 0 (i.e., pruned weights trivially satisfy that their effect on forward activations cannot depend on width or sparsity). We can rewrite Equation \ref{eqn:weight-update-lln} as:
\begin{align}
    \E[\Delta \YB_{ij}] \to \eta m_{d_\text{in}}d_\text{in,base}m_\rho \rho_\text{base} \E \left[ \XB_{ik} \left( \frac{\sum^T_t \gamma_t \sum_b^B \XB^{t}_{bk} \nabla_{\YB} \mathcal{L}^{t}_{bj} }{\sqrt{\sum_t^T \omega_t \sum_b^B (\XB^{t}_{bk} \nabla_{\YB} \mathcal{L}^{t}_{bj})^2}} \right) \right], \text{ as } (d_\text{in}\rho) \to \infty
\end{align}

\textbf{Solution:} For Adam updates, to satisfy the FLD and ensure $\E[\Delta \YB_{ij}]$ and the typical entry size of $\Delta \YB$ are scale invariant to $m_{d_\text{in}}$ and $m_{\rho}$, we choose $\eta = \frac{\eta_{\text{base}}}{m_{d_\text{in}}m_{\rho}}$. Note that this correction is equivalent to \textmu P \cite{yang2022mup} when $\rho=1,m_\rho=1$.

\subsection{Additional notes about derivation}

We make a few supplementary notes about the above derivation:
\begin{itemize}
    \item Throughout our derivation, we use $\rho$ to refer to the density level. Note that since this derivation is local to a single layer in the model, the density (or sparsity) level can also be parameterized independently for each layer. If a sparsity technique will use layer-wise independent sparsity levels, appropriate corrections should be made for each layer.
    \item Similar to the $\rho$ notation, we use $\eta$ to denote the learning rate, but this learning rate can be layer-specific depending on sparsity level. Appropriate corrections must be made if using layer-wise independent sparsities.
    \item The use of the Law of Large Numbers in portions of the above derivation indicate that \ac{smup} is expected to provide stable training dynamics as the number of non-zero weights per neuron (WPN) tends to infinity. However, in sparse settings, the WPN can tend to be small. If WPN is small, training dynamics may be affected, and this might be a direction for future work.
    \item In this work, we only consider sparsifying linear projection layers. As a result, \ac{smup} matches \ac{mup} for other layers like input, output, bias, layer-norm, and attention logits. Depending on the sparsification technique, these other layers might need to be reviewed for their effects on training dynamics.
\end{itemize}

\section{Experimental details} \label{sec:experiment-details}

\paragraph{\ac{smup} base hyperparameter tuning} To find the optimized set of hyperparameters for \ac{smup}, we actually tune \ac{mup} HPs on a dense proxy model. By formulation of \ac{smup}, these HPs transfer optimally to all the sparse models trained for this work. This dense proxy model is a GPT-2 model, but with small changes: ALiBi position embeddings~\cite{press2022alibi} and SwiGLU nonlinearity~\cite{shazeer2020glu}. We configure it with width: $d_{\text{model}} = d_{\text{model,base}} = 256$, number of layers: $n_{\text{layers}} = 24$, and head size: $d_{\text{head}} = 64$, resulting in 39M parameters. We trained this proxy model on 800M tokens with a batch size of 256 sequences and sequence length 2048 tokens. We randomly sampled 350 configurations of base learning rates, base initialization standard deviation, and embedding and output logits scaling factors. From this sweep we obtained the tuned hyperparameters listed in Table~\ref{tab:proxy-model-hps}.

\begin{table}[h]
    \centering
    \caption{Tuned hyperparameters for our dense proxy model.}
    \begin{tabular}{cc}
         \toprule
         Hyperparameter& Value\\
         \midrule
         $\sigma_{W,\text{base}}$& 0.08665602\\
         $\eta_{\text{base}}$& 1.62E-2\\
         $\alpha_{\text{input}}$& 9.1705\\
         $\alpha_{\text{output}}$& 1.0951835\\
         \bottomrule
    \end{tabular}
    \label{tab:proxy-model-hps}
\end{table}

\paragraph{Experimental details for all figures}
In Table \ref{tab:experimental-details}, we provide extensive details on hyperparameters, model size, and training schedule for all experiments in this paper. All models in this paper were trained on the SlimPajama dataset~\cite{cerebras2023slimpajama}, a cleaned and deduplicated version of the RedPajama dataset.

\begin{sidewaystable}[]
\centering
\small
\caption{Experimental details for all figures in this paper.}
\begin{tabular} {lrrrrrrrrlrrr}
\toprule
& & & & & & & & & & LR warm- & & \\
Figure & $d_{\text{model}}$ & $L$ & $d_{\text{head}}$ & $B$ & LR & Init. Stdev. & $\alpha_{\text{input}}$ & $\alpha_{\text{output}}$ & LR decay & up steps & Steps & Tokens \\
\hline
\hline
Fig. \ref{fig:stable_hps}, & 4096 & 2 & 64 & 128 & Variable & SP: 2.166E-2 & 9.1705   & 1.095  & 10x linear & 116& 1169  & 306M \\
\ref{fig:mutransfer-lr}, \ref{fig:smup_pareto_frontier}, \ref{fig:mutransfer-ablation-lr} & & & & & & \ac{mup}, \ac{smup}: 0.087 & & & & & & \\
\hline
Fig. \ref{fig:mutransfer-init}, \ref{fig:mutransfer-ablation-init} & 4096 & 2 & 64 & 8 & 1.62E-2 & Variable & 9.1705   & 1.095  & Constant & 0 & 100  & 1.6M \\
\hline
Fig. \ref{fig:demo-reduced-dense-20tpp} & 2048 & 10 & 64 & 504 & SP: 2e-4 & SP: 0.02 & 9.1705 & 1.095  & Decay to zero & 1175 & 11752 & 12.13B \\
& & & & & \ac{mup}, \ac{smup}: 1.62E-2 & \ac{mup}, \ac{smup}: 0.087 & & & & & & \\
\hline
Fig. \ref{fig:coord-check} & 2048 & 2 & 32         & 4   & 1.68E-02         & 0.101           & 11.22    & 1            & Constant         & 0  & 10    & 82K    \\
& & & & & \ac{mup}, \ac{smup}: 1.62E-2 & & & & & & & \\
\hline
Fig. \ref{fig:mutransfer-lr-iso-flop}     & Variable    & 2   & 64         & 128 & Variable     & SP: 0.02 & 9.1705   & 1.095  & 10x linear & 116& 1169  & 306M   \\
 & & & & & & \ac{mup}, \ac{smup}: 0.087 & & & & & & \\
\hline
Fig. \ref{fig:demo-iso-flop} & Variable    & 10  & 64         & 256 & SP, $d_{\text{model}} \le$ 1088: 6E-4 & SP: 0.02 & 9.1705   & 1.095  & 10x linear & 190 & 1907  & 1B     \\
& & & & & SP, $d_{\text{model}} >$ 1088: 2E-4 & \ac{mup}, \ac{smup}: 0.087 & & & & & &     \\
& & & & & \ac{mup}, \ac{smup}: 1.62E-2 & & & & & & &     \\
\hline
Fig. \ref{fig:mutransfer-lr-iso-wpn}      & Variable    & 2   & 64         & 128 & Variable     & 0.087 for SP.          & N/A      & N/A          & 10x linear & 116& 1169  & 306M   \\
\hline
Fig. \ref{fig:dst-mutransfer-lr} & 1024 & 2 & 64 & 128 & Variable & SP: 2.166E-2 & 9.1705   & 1.095  & 10x linear & 116 & 1169  & 306M \\
 & & & & & & \ac{mup}, \ac{smup}: 0.087 & & & & & & \\

\bottomrule
\end{tabular}
\label{tab:experimental-details}
\end{sidewaystable}

\end{document}